\documentclass[dvipsnames]{valence} 

\usepackage{soul} 
\usepackage{float}
\usepackage{amsmath}
\usepackage{algorithm}
\usepackage{algpseudocode}
\usepackage{amsfonts}
\usepackage{subcaption}
\usepackage{longtable}

\usepackage[export]{adjustbox}
\usepackage{comment}
\usepackage{mathtools}
\usepackage{authblk}
\usepackage{fancyhdr}
\usepackage{multirow}
\usepackage{booktabs}
\usepackage{array}
\usepackage{hyperref}
\usepackage[capitalize,noabbrev]{cleveref}
\usepackage{lipsum}
\usepackage{xspace}

\crefname{boxcounter}{Box}{Boxes}

\title{Towards Autonomous Mechanistic Reasoning in Virtual Cells}

\setrunningtitle{Towards Autonomous Mechanistic Reasoning in Virtual Cells}

\author[1 \star]{Yunhui Jang}
\author[2, 3]{Lu Zhu}
\author[4 \star]{Jake Fawkes}
\author[2,3]{Alisandra Kaye Denton}
\author[2,3]{Dominique Beaini}
\author[2,3,\dagger]{Emmanuel Noutahi}

\affiliation[1]{Korea Advanced Institute of Science and Technology (KAIST)}
\affiliation[2]{Valence Labs}
\affiliation[3]{Recursion}
\affiliation[4]{University College London}
\contribution[\star]{\small Work done during an internship at Valence Labs}
\contribution[\dagger]{\small Correspondence: emmanuel@valencelabs.com}

\newcommand{\Methodname}{$\textsc{VCR-Agent}$\xspace}
\newcommand{\datasetname}{$\textsc{VC-Traces}$\xspace}

\definecolor{myblue}{HTML}{1436BC}
\definecolor{mylightblue}{HTML}{87B7D8}
\definecolor{mypurple}{HTML}{B046E6}
\definecolor{mybrown}{HTML}{D55E00} 

\abstract{
Large language models (LLMs) have recently gained significant attention as a promising approach to accelerate scientific discovery. However, their application in open-ended scientific domains such as biology remains limited, primarily due to the lack of factually grounded and actionable explanations. To address this, we introduce a structured explanation formalism for virtual cells that represents biological reasoning as mechanistic action graphs, enabling systematic verification and falsification. Building upon this, we propose \Methodname, a multi-agent framework that integrates biologically grounded knowledge retrieval with a verifier-based filtering approach to generate and validate mechanistic reasoning autonomously. Using this framework, we release \datasetname dataset, which consists of verified mechanistic explanations derived from the Tahoe-100M atlas. Empirically, we demonstrate that training with these explanations improves factual precision and provides a more effective supervision signal for downstream gene expression prediction. These results underscore the importance of reliable mechanistic reasoning for virtual cells, achieved through the synergy of multi-agent and rigorous verification.
}

\begin{document}

\maketitle

\section{Introduction}

The development of virtual cells, computational models that simulate cellular behavior, promises to advance biological discovery and drug design~\citep{noutahi2025virtualcellspredictexplain,bunne2024build, adduri2025predicting}. A central goal of these models is to accurately predict cellular responses to perturbations, such as genetic knockouts or drug treatments. To move beyond correlation-based prediction toward actionable insight, virtual cells must also produce mechanistically grounded explanations. However, generating such explanations that are both biologically plausible and reliable remains a critical bottleneck. Large language models (LLMs) have emerged as a potential solution, demonstrating strong reasoning capabilities in domains such as mathematics and programming~\citep{shao2024deepseekmathpushinglimitsmathematical, chen2021evaluatinglargelanguagemodels, deepseekai2025deepseekr1incentivizingreasoningcapability, openai2024openaio1card}. These capabilities are acquired by training on vast, high-quality reasoning datasets.

However, directly transferring the reasoning training paradigms from mathematics or coding to scientific discovery is not straightforward. From a data-centric perspective, a key difficulty lies in the curation of large-scale, reliable reasoning datasets. The datasets that power LLM reasoning are typically derived from two primary sources: (1) high-quality human annotations~\citep{cobbe2021gsm8k, gao2024omni, hendrycksmath2021}, or (2) large-scale synthetic generation by LLMs~\citep{wang-etal-2023-self-instruct, moshkov2025aimo2, guha2025openthoughtsdatarecipesreasoning}. Human annotation, while high-quality, is prohibitively expensive and non-scalable in domains that require specialized expertise, such as biology. Conversely, LLM-generated reasoning traces are often factually unreliable and prone to hallucination, particularly in settings where LLMs lack sufficient domain grounding. This limits their applicability in scientific contexts where factual correctness and reliability are essential.

Beyond data scarcity, a more fundamental challenge lies in the verification of reasoning. In mathematics and programming, reasoning traces can be automatically verified for correctness, e.g., code can be executed, and its output is checked against a ground truth. Biological reasoning, in contrast, rarely admits such direct verification because it relies on disjointed knowledge from scientific literature rather than deterministic rules. This inherent ambiguity makes it difficult to verify the correctness of a reasoning trace and constitutes a critical bottleneck for the reliable use of LLMs in scientific explanation.

\begin{figure*}[t]
    \centering
    \includegraphics[width=0.97\linewidth]{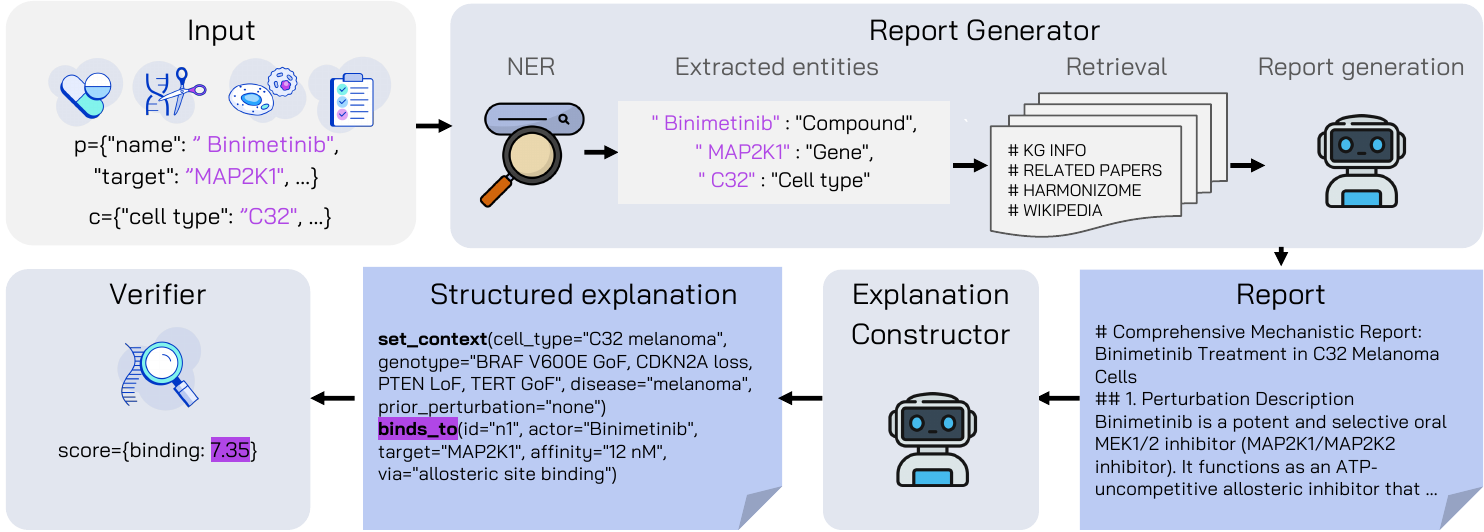}
\caption{\textbf{An Overview of the \Methodname Multi-Agent Framework.} The \textit{Report Generator} accepts the perturbation and cellular context, performing knowledge retrieval and synthesis to produce a comprehensive, biologically grounded report. The \textit{Explanation Constructor} then translates this report into the formal structured mechanistic explanation. This generated structured explanation is subsequently evaluated by the \textit{Verifier} for factual validation and filtering.} \label{fig3:main}
\vspace{-0.2in}
\end{figure*}

To overcome these challenges, we introduce \emph{structured explanations} for virtual cells, which constrain biological reasoning into explicit mechanistic actions connected by directed dependencies. Intuitively, rather than relying on ambiguous free-form natural language, we treat the explanation as a sequence of discrete, biologically-grounded actions. Each action consists of a predefined primitive and domain-specific arguments spanning molecular interactions to phenotype manifestations. By explicitly encoding biological dependencies, such as preconditions or regulatory requirements, these actions are logically connected to form a directed graph. In this representation, nodes constitute the discrete actions, while edges define the mechanistic relationships and dependencies between them. This format effectively restricts the model to a predefined action space, ensuring that every generated explanation is both interpretable and falsifiable. By grounding the reasoning in established biology, our framework produces \emph{mechanistically plausible structures} that provide a rigorous basis for hypothesis generation, while remaining distinct from formal interventional causal discovery.

Building on this formalism, we propose \Methodname, a multi-agent system that orchestrates the generation and validation of structured mechanistic reasoning, as illustrated in \cref{fig3:main}. Our system decomposes the reasoning process into two specialized modules: (1) a \textit{report generator} that aggregates and summarizes factual biological knowledge from external databases, thereby resolving the issue of factual grounding; and (2) an \textit{explanation constructor} that transforms this retrieved report into the proposed structured format, mitigating the ambiguity of unstructured and unverifiable reasoning. To guarantee scientific reliability, each generated explanation undergoes \textit{verifier-based filtering}, where specialized verifiers evaluate the explanation traces to retain only those that are factually accurate and causally coherent.

We apply our framework to the Tahoe-100M~\citep{Zhang2025.02.20.639398} atlas to generate and publicly release a \datasetname dataset of structured mechanistic explanation traces. We evaluate this dataset through explanation quality assessment and downstream gene expression prediction under perturbation. Our experiments show that models trained on these verified reasoning traces achieve stronger downstream performance compared to baselines.

We summarize our contributions as follows:

\begin{itemize}
    \item We define a structured explanation format for virtual cells that supports interpretability and falsifiability through biology-grounded verifiers.
    \item We propose \Methodname, a multi-agent system that integrates knowledge retrieval, structured reasoning generation, and a verifier-based filtering to ensure biological reliability of generated traces.
    \item We release \datasetname, a dataset of structured explanations to facilitate research in virtual cell reasoning.
    \item We empirically demonstrate that our verified structured explanations are high-quality and improve the performance of LLMs on gene-related downstream tasks, underscoring their practical application.
\end{itemize}

\section{Structured Mechanistic Reasoning for Virtual Cells}\label{sec:2_structured_reasoning}

\begin{figure*}
    \centering
     \begin{subfigure}[t]{0.65\textwidth}
    \includegraphics[width=\linewidth]{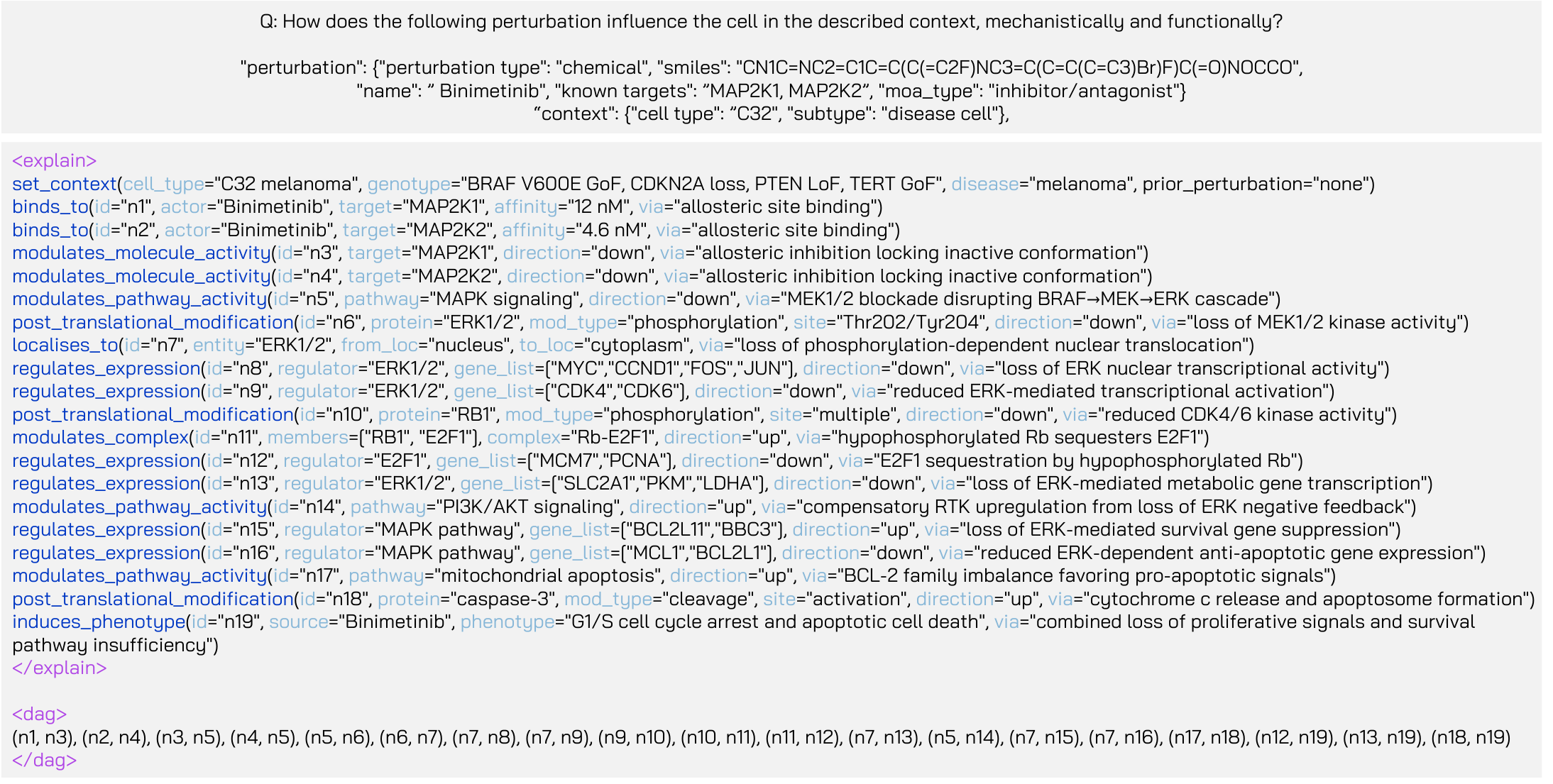}
    \caption{\textbf{An example of mechanistic reasoning traces.}}
    \label{fig2_1:reasoning_example}
    \end{subfigure}
    \centering
    \begin{subfigure}[t]{0.34\textwidth}
        \includegraphics[width=\linewidth]{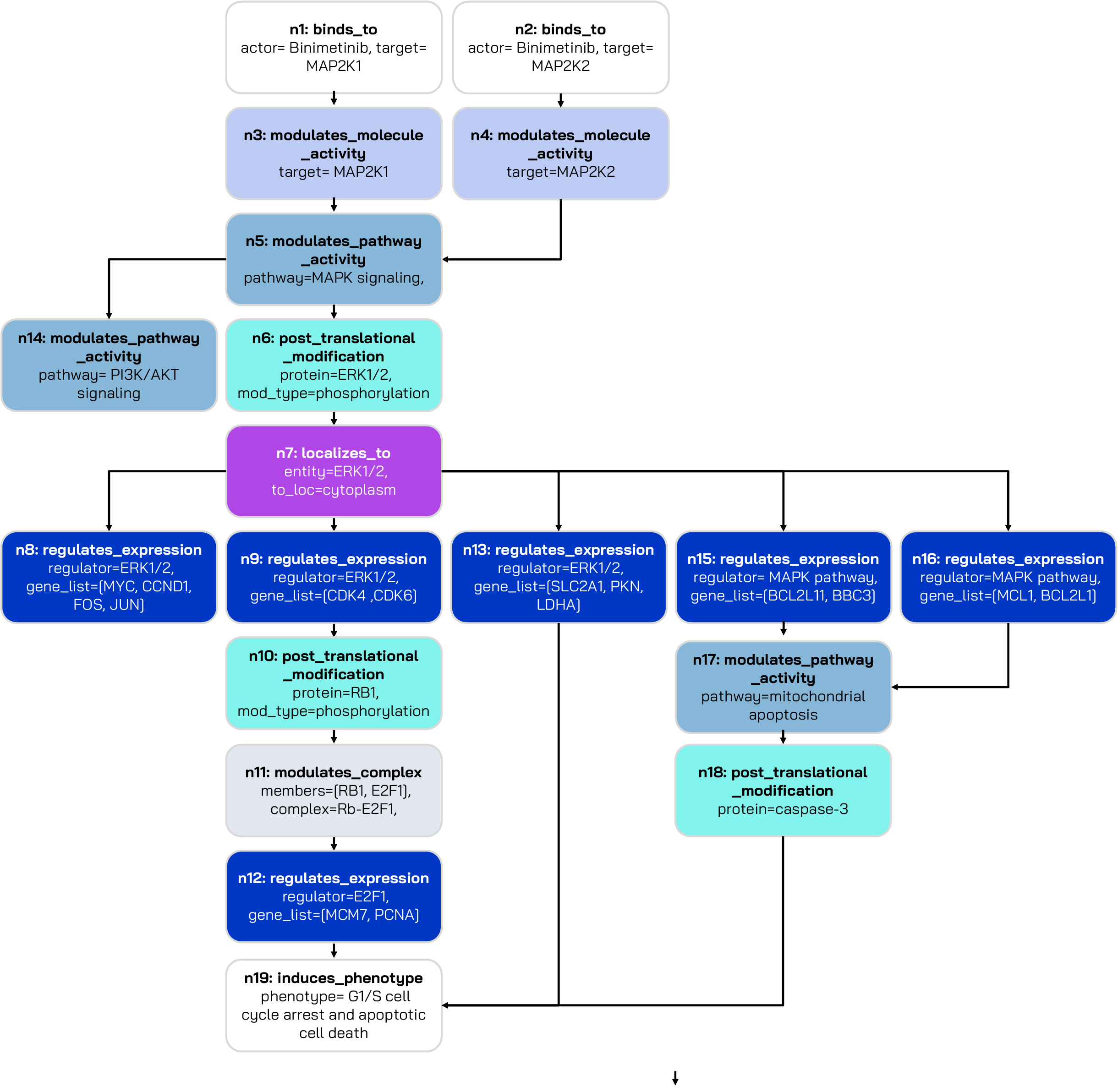}
        \captionsetup{justification=centering}
        \caption{\textbf{An example of DAG.}}\label{fig2_2:dag}
    \end{subfigure}
    \caption{\textbf{An overview of structured reasoning.} (\subref{fig2_1:reasoning_example}) Given an input $(p,c)=$(Binimetinib, C32), the model generates mechanistic reasoning traces. \textcolor{myblue}{Blue} and \textcolor{mylightblue}{light blue} indicate the action primitives and the arguments, respectively. The elements within the \texttt{<dag>} tag represent the edge list defining the reasoning graph. (\subref{fig2_2:dag}) An example of DAG. Same color indicates the same action primitive.} \label{fig2:structured_reasoning}
    \vspace{-0.2in}
\end{figure*}

Virtual cells must be capable of structured, autonomous reasoning that explains how molecular perturbations lead to observable cellular outcomes. To achieve this, we adopt a structured formulation that serves two primary purposes: (1) it constrains the reasoning space to ensure interpretability and faithfulness, and (2) it enables automatic falsification via biology-grounded verifiers. To align such reasoning processes with the inherent topology of cellular signaling, where information propagates through cascades of mechanistic events, we formalize structured reasoning as the task of inferring a directed acyclic graph (DAG) of mechanistic interactions given a perturbation context. An overview of this problem formulation and an example mechanistic reasoning trace are illustrated in \cref{fig2_1:reasoning_example}, while the DAG structure is visualized in \cref{fig2_2:dag}.

\subsection{Problem Formulation}

The goal of structured reasoning in virtual cells is to infer how a given perturbation affects the cellular state through a series of mechanistic steps. Formally, given an input $x = (p, c)$, where $p$ denotes the \textit{perturbation} (e.g., a chemical compound, genetic knockdown, etc.) and $c$ denotes the \textit{cellular context} (e.g., cell type, disease model, etc.), we aim to generate a reasoning graph $\mathcal{G}$ that captures a chain of mechanistic actions triggered by the perturbation $p$ in the context $c$.

The output reasoning is represented as a DAG $\mathcal{G}$:
\begin{equation*}
    \mathcal{G} = (\mathcal{V}, \mathcal{E}), \quad \text{where } \mathcal{V} = \{ n_1, \dots, n_k \}.
\end{equation*}

Each node $n_i \in \mathcal{A}$ represents a mechanistic action such as binding, modulation, regulation, etc., comprising an action primitive selected from the predefined action space $\mathcal{A}$ (defined in \cref{subsec:2_2_action_space}) and its associated arguments. Each directed edge $(n_i, n_j) \in \mathcal{E}$ represents a mechanistic dependency, indicating that the outcome of action $n_i$ enables or influences another action $n_j$. For example, a ligand–receptor binding (\texttt{binds\_to}) may precede a downstream signaling modulation (\texttt{modulates\_pathway\_activity}).

Finally, the reasoning model $f_\theta$ is defined as
\begin{equation*}
f_\theta: x \rightarrow \mathcal{G},
\end{equation*}
where $f_\theta$ generates both the mechanistic actions (nodes) and their dependencies (edges). This representation encodes mechanistic plausibility and downstream biological consequences, thereby enhancing the interpretability of the reasoning model's logic while remaining distinct from formal, interventional causal discovery.

\subsection{Action Spaces}\label{subsec:2_2_action_space}

The \textit{action space} $\mathcal{A}$ defines the set of permissible reasoning actions for a virtual cell. Constraining each reasoning step to a finite and biologically grounded set of high-confidence primitives enables falsifiability, as verifiers can evaluate actions. To ensure systematic consistency, each primitive is parameterized by a specific argument schema, such as assigning an actor and target to a \texttt{binds\_to} action.

We define twenty action primitives grouped into seven categories: (1) system initialization, (2) metabolic, (3) regulation, (4) functional, (5) interaction, (6) phenotype, and (7) proteostasis. These categories span from molecular interactions to phenotypic manifestations. We provide an overview of defined action primitives in \cref{fig2_2:action_spaces} and detailed argument schemas in \cref{appx: details_action}.

To illustrate how an action primitive is defined and parameterized, consider an example action \texttt{binds\_to}. This action specifies a direct molecular interaction between two biomolecules, such as a drug–target, ligand–receptor, or protein–protein pair. It is parameterized as:

\begin{figure}
    \centering
    \includegraphics[width=0.5\linewidth]{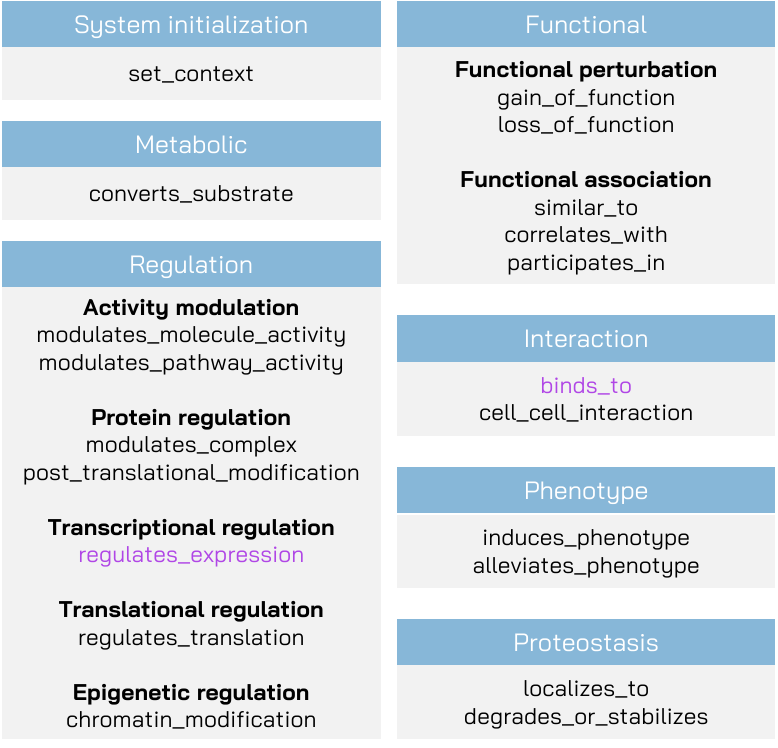}
    \caption{\textbf{An overview of action spaces.} The sub-categories are represented with \textbf{bold} and action primitives with verifier are represented with \textcolor{mypurple}{purple}. The argument schemes are in \cref{appx: details_action}.}  \label{fig2_2:action_spaces}
    \vspace{-0.2in}
\end{figure}

\[
\begin{aligned}
\texttt{binds\_to}(&\textit{id, actor, target, \{affinity, unit, residues\_actor, residues\_target, via, confidence\}}).
\end{aligned}
\]

The \textit{id} argument specifies the node identifier used to connect the actions in the DAG, while \textit{actor} and \textit{target} define the participating entities. Other optional arguments represented in \{ \} are mapped to biological ontologies such as compounds, proteins, affinity scores, etc. By leveraging these structured arguments, verifiers can evaluate the reliability of each action with automatic verification against curated databases~\citep{mendez2019chembl,ashburner2000gene,croft2010reactome} and computational tools~\citep{Passaro2025.06.14.659707,love2014moderated}.

\section{LLM-Agent Framework for Reasoning}\label{sec:3_llm_agent}

We introduce \Methodname, our multi-agent system designed to generate structured explanations for virtual cells given input perturbations and cellular contexts. The framework is designed as a two-stage pipeline to ensure factual grounding and structured output, consisting of a \textit{report generator} and an \textit{explanation constructor}. We provide an overview in \cref{fig3:main}.

First, the report generator is responsible for information retrieval and summarization. It queries comprehensive knowledge bases to gather relevant biological facts about the given perturbation and cellular context. This is then summarized into a comprehensive natural-language report.

Next, the explanation constructor takes this knowledge-grounded report as its input, transforming it into the structured reasoning format. By decoupling knowledge acquisition from structured reasoning generation, this enforces knowledge grounding, thereby ensuring both the factual accuracy and structural integrity of the final explanation.

\subsection{Report Generator} 

\begin{figure}
    \centering
    \includegraphics[width=0.5\linewidth]{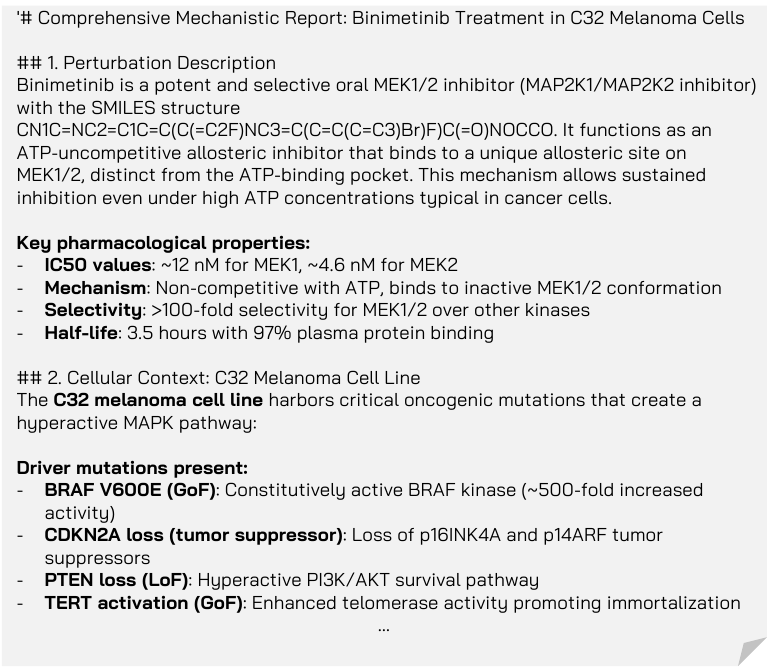}
    \caption{\textbf{An example of generated report.} The input perturbation - cellular context pair follows the one in \cref{fig2_1:reasoning_example}. } \label{fig3_2:report_generator}
    \vspace{-0.2in}
\end{figure}

The report generator operates in a three-step process: (1) entity extraction with biomedical name entity recognition (NER), (2) retrieval with external knowledge bases, and (3) report generation based on the related retrieved information.

First, the entity extraction step identifies relevant biomedical entities from the given input perturbation and cellular context. We employ HunFlair2~\citep{sanger2024hunflair2}, which extracts the biomedical entities, including chemical compounds, genes, diseases, etc. This NER process simplifies the retrieval by enabling entity-based search instead of relying on complex natural language queries.

Next, the knowledge retrieval step uses these extracted entities to query external knowledge bases to aggregate relevant biological facts. Specifically, these include StarkPrimeKG, a biomedical knowledge graph~\citep{wu24stark}; Harmonizome, a gene-related database~\citep{10.1093/nar/gkae1080}; PubMed, a biomedical literature database; and Wikipedia.

The retrieval process operates as follows:
\begin{itemize}
    \item \textbf{StarkPrimeKG:} We query the entity and its synonyms to search for the matching node in the knowledge graph. If no exact match is found, we identify the most similar node based on the cosine similarity of PubMedBERT~\citep{gu2021domain} embeddings. The 1-hop neighbor nodes are then aggregated into a textual context to provide relevant relational information.
    \item \textbf{Harmonizome:} We query gene entities to enrich gene-specific information. These entities include both the genes encoding protein targets of compound perturbations and the genes targeted by genetic perturbations.
    \item \textbf{PubMed:} We query the database using a set of extracted entities to identify relevant literature, prioritizing papers whose abstracts demonstrate the highest similarity to the input entities.
    \item \textbf{Wikipedia:} We query each entity to retrieve the best matching documents from Wikipedia.
\end{itemize}

We provide the examples of information retrieved from four databases in \cref{appx: example}.

Finally, the report generation step summarizes all retrieved information into a single, comprehensive report. We prompt Claude 4~\citep{claude4} with the retrieved information to generate the report. This report provides comprehensive information on the input perturbation and cellular context, which serves as the factual foundation for the explanation constructor. We provide an example generated report in \cref{fig3_2:report_generator} and the used prompt in \cref{appx: exp}.

\subsection{Explanation Constructor}

Next, the explanation constructor generates the structured reasoning, based on the knowledge-grounded report as input. This process enables the verification and falsification of the explanation. We also used Claude 4~\citep{claude4} in the same way as the report generation step, and the prompt used for explanation generation is provided in \cref{appx: exp}.
\section{Verifier-based Filtering and Quality Control}\label{sec:4_verifier}

To ensure the biological and factual accuracy of the explanations generated by \Methodname, we introduce a verifier-based filtering and quality control pipeline. This is critical for mitigating hallucinations and pruning reasoning traces that conflict with biological facts. The process consists of two stages: per-action verification and filtering.

First, individual actions within a generated explanation trace are evaluated by a corresponding, specialized verifier. Our framework supports diverse verifiers tailored to various action primitives. In this study, we implement four verifiers spanning the most frequently occurring action types: drug-target interaction (DTI), differential expression (DE), {subcellular localization (LOC), and phenotype (PHENO) verifiers. Among these, DTI and DE serve as the primary verifiers used for filtering and are most directly relevant to the downstream TahoeQA task (\cref{sec5_3:tahoeqa}).}

Next, the resulting verification scores are utilized to filter out either the entire explanation trace or partial arguments that fail to meet the plausibility thresholds. This rigorous, multi-level validation ensures that the final explanation traces are both factually accurate and logically coherent.

\begin{figure}
    \centering
    \includegraphics[width=0.5\linewidth]{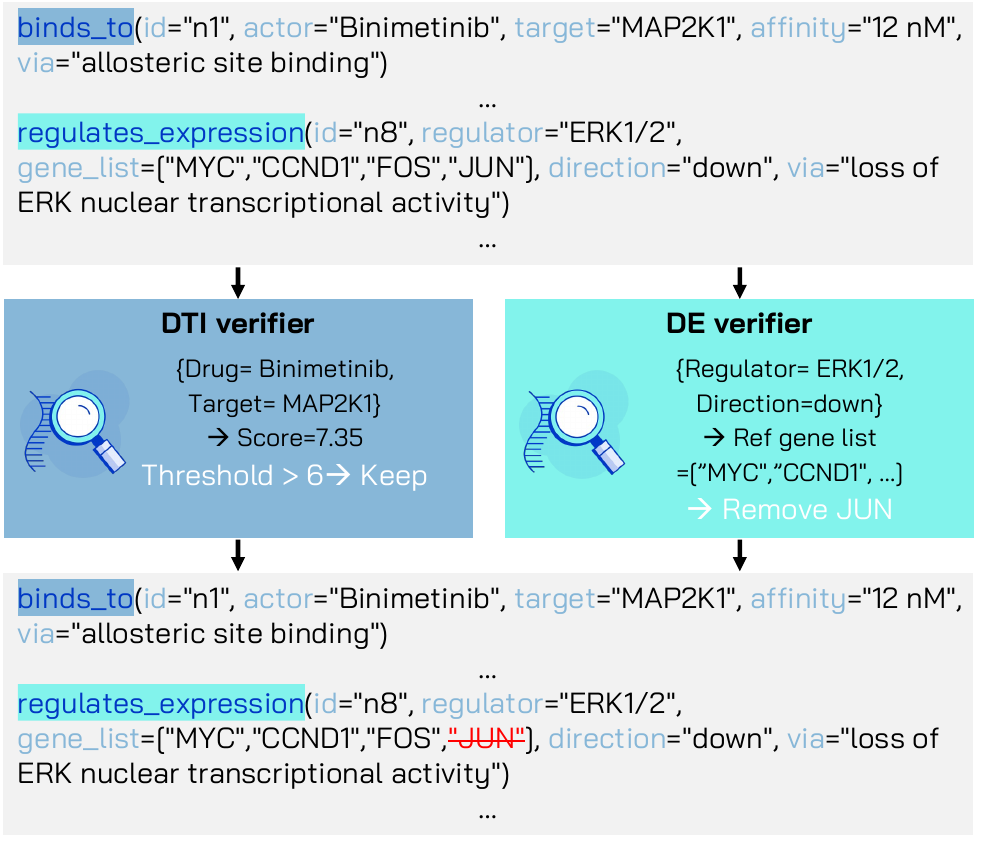}
    \caption{\textbf{An example of verifier-based filtering process.} The pipeline processes initial structured explanation (top) through verifiers (middle) to produce filtered output (bottom). Same colors link the action primitive to their corresponding verifiers.} \label{fig4_2:verifier_filtering}
    \vspace{-0.1in}
\end{figure}

\subsection{Verifier}\label{subsec4_1:verifiers}

Here, we introduce our biologically specialized verifiers, designed to quantify the validity of specific actions or arguments within generated explanations and identify potential hallucinations. We detail the two primary verifiers used for filtering: a DTI verifier for the \texttt{binds\_to} action and a DE verifier for the \texttt{regulates\_expression} action. {We note that additional verifiers are described in \cref{appx: additional_verifiers}.}

First, the DTI verifier predicts the physical plausibility of a binding between a given actor (drug) and target (protein). It leverages Boltz-2~\citep{Passaro2025.06.14.659707} to model the protein-ligand interaction and characterize the binding interface. This process yields a continuous binding probability score.

Next, the DE verifier validates whether the perturbation up- or down-regulates the target gene. This verifier queries ground-truth differential expression datasets, such as \mbox{Tahoe-100M}~\citep{Zhang2025.02.20.639398}, to confirm if the predicted target gene is significantly regulated in the given direction. This identifies hallucinated gene targets, flagging predictions that contradict established biological knowledge.

\begin{table*}[t]
    \centering
    \caption{\textbf{Explanation quality performance.} The best results are highlighted in \textbf{bold}. The standard deviation is computed across cell lines.}\label{tab_5_1:main}
    \resizebox{0.85\linewidth}{!}{
    \begin{tabular}{cccccccc}
    \toprule[1.25pt]
    && \multicolumn{2}{c}{Format} & \multicolumn{2}{c}{Verifier} \\
        \cmidrule(lr){3-4}  \cmidrule(lr){5-6}
        \multicolumn{2}{c}{\textbf{Model}} & Validity & Verifiability &  Drug-target interaction & Differential expression \\
        \midrule
         \multirow{3}{*}{\textbf{Open}} & DeepSeek-R1-8B & 0.003 \scriptsize{$\pm$ 0.004} & 1.000 \scriptsize{$\pm$ 0.081} & 0.000 \scriptsize{$\pm$ 0.000} & 0.000 \scriptsize{$\pm$ 0.000} \\
         & Qwen3-30B & 0.860 \scriptsize{$\pm$ 0.016} & 0.965 \scriptsize{$\pm$ 0.007} & 0.528 \scriptsize{$\pm$ 0.004} & 0.272 \scriptsize{$\pm$ 0.030} \\
         & Llama3.3-70B & 0.841 \scriptsize{$\pm$ 0.046} & 0.887 \scriptsize{$\pm$ 0.017} & 0.322 \scriptsize{$\pm$ 0.034} & 0.090 \scriptsize{$\pm$ 0.010} \\
         \midrule
         \multirow{2}{*}{\textbf{Closed}} & Claude-Sonnet-4 & \textbf{1.000} \scriptsize{$\pm$ 0.000} & 0.867 \scriptsize{$\pm$ 0.008} & 0.657 \scriptsize{$\pm$ 0.005} & 0.504 \scriptsize{$\pm$ 0.064} \\
         & \Methodname (Ours) & \textbf{1.000} \scriptsize{$\pm$ 0.000} & \textbf{0.945} \scriptsize{$\pm$ 0.011} & \textbf{0.725} \scriptsize{$\pm$ 0.017} & \textbf{0.528} \scriptsize{$\pm$ 0.060} \\ 
     \bottomrule[1.25pt]
    \end{tabular}}
    \vspace{-0.2in}
\end{table*}

\subsection{Verifier-based Filtering}\label{subsec4_2:filtering}
To guarantee that the generated explanation traces are biologically grounded, we apply a filtering process based on the two verifiers defined in \cref{subsec4_1:verifiers}. {These two action types jointly appear in 91.5\% of all explanation traces in \datasetname, these verifiers provide broad filtering coverage in practice.}

First, we enforce a validity constraint on molecular interactions; specifically, any trace containing a \texttt{binds\_to} action with a DTI confidence score below a pre-defined threshold $\tau$ is discarded. Second, we employ the DE verifier to prune factually inconsistent predictions. This step eliminates gene arguments that correspond to incorrectly identified or directionally mismatched gene expression changes. {Notably, our filtering is designed to prevent false positives, i.e., it only removes claims that directly contradict established biological evidence, leaving others unchanged.} We provide an illustrative example of this verifier-based filtering in \cref{fig4_2:verifier_filtering}, depicting a scenario where suboptimal explanation traces are identified and flagged for verification.

\section{Experiments}

In this section, we evaluate both the effectiveness of the \Methodname framework and the quality of the resulting \datasetname dataset. Our evaluation is structured into two primary components: (1) an evaluation of explanation quality, and (2) an evaluation of the dataset's utility as a supervision signal for downstream biological tasks. We provide detailed experimental settings in \cref{appx: exp} and an additional ablation study in \cref{appx:additional_exp}.

\subsection{Explanation Quality}\label{sec:5_1_explanation_quality}

We first assess the quality of the structured explanations generated by our \Methodname.

\paragraph{Dataset.} We derived our experimental dataset from the Tahoe-100M atlas~\citep{Zhang2025.02.20.639398}, extracting a total of 18,950 unique compound perturbation-context pairs. Using these pairs, we constructed the \datasetname dataset through the \Methodname framework, which transforms the perturbations into mechanistic reasoning traces. To align with the test split of Tahoe-X1~\citep{gandhi2025tahoe}, we focus our experiment on a representative subset of this dataset comprising five cell lines (C32, HOP62, HepG2/C3A, Hs 766T, and PANC-1). Notably, our complete \datasetname dataset is publicly released in {\href{https://github.com/valence-labs/VC-TRACE}{https://github.com/valence-labs/VC-TRACE}}.

\paragraph{Baselines.} We compare \Methodname with both open-source and closed-source LLMs. For open-source baselines, we use three models: Qwen3-30B-A3B~\citep{yang2025qwen3technicalreport}, DeepSeek-R1-0528-Qwen3-8B~\citep{deepseekai2025deepseekr1incentivizingreasoningcapability}, and Llama3.3-70B-Instruct~\citep{grattafiori2024llama}. For the closed-source baseline, we use Claude-Sonnet-4~\citep{claude4}, the same base model used within \Methodname for fair comparison.

\paragraph{Metrics.} We evaluate performance using two format-based metrics and the two verifier scores. First, regarding format, we report validity, which measures the proportion of traces that are both syntactically correct, i.e., containing proper \texttt{<explain>} and \texttt{<dag>} tags, and structurally valid, meaning all generated action primitives adhere to the definitions in \cref{subsec:2_2_action_space}. We also report verifiability, which quantifies the proportion of generated arguments that can be successfully mapped to valid biomedical entities for verification. Finally, the verifier scores are computed as described in \cref{sec:4_verifier} and averaged to assess the factual correctness of individual mechanistic actions. Notably, the DTI score is computed as the average of the binding score, while the DE score is computed as the proportion of the traces where at least a single DE step included in the trace is correct.

\paragraph{Results.} We present the results in \cref{tab_5_1:main}. We observe that \Methodname consistently generates high-quality reasoning traces across both format and verifier-based metrics. These findings demonstrate that our framework generates explanations that are not only structurally valid and verifiable but also more closely aligned with biological reference data than those produced by the baselines.

Crucially, the results reported in \cref{tab_5_1:main} evaluate the raw generative performance of the models prior to any filtering. While these metrics already demonstrate superior alignment with biological verifiers compared to baselines, our framework further enhances reliability through the verification pipeline: during the construction of \datasetname, this process successfully excluded 28.2\% of faulty DTI claims and refined 87.3\% of DE actions to eliminate hallucinations.

\subsection{Application: TahoeQA}\label{sec5_3:tahoeqa}

\begin{figure*}[t]
    \centering
    \includegraphics[width=0.98\linewidth]{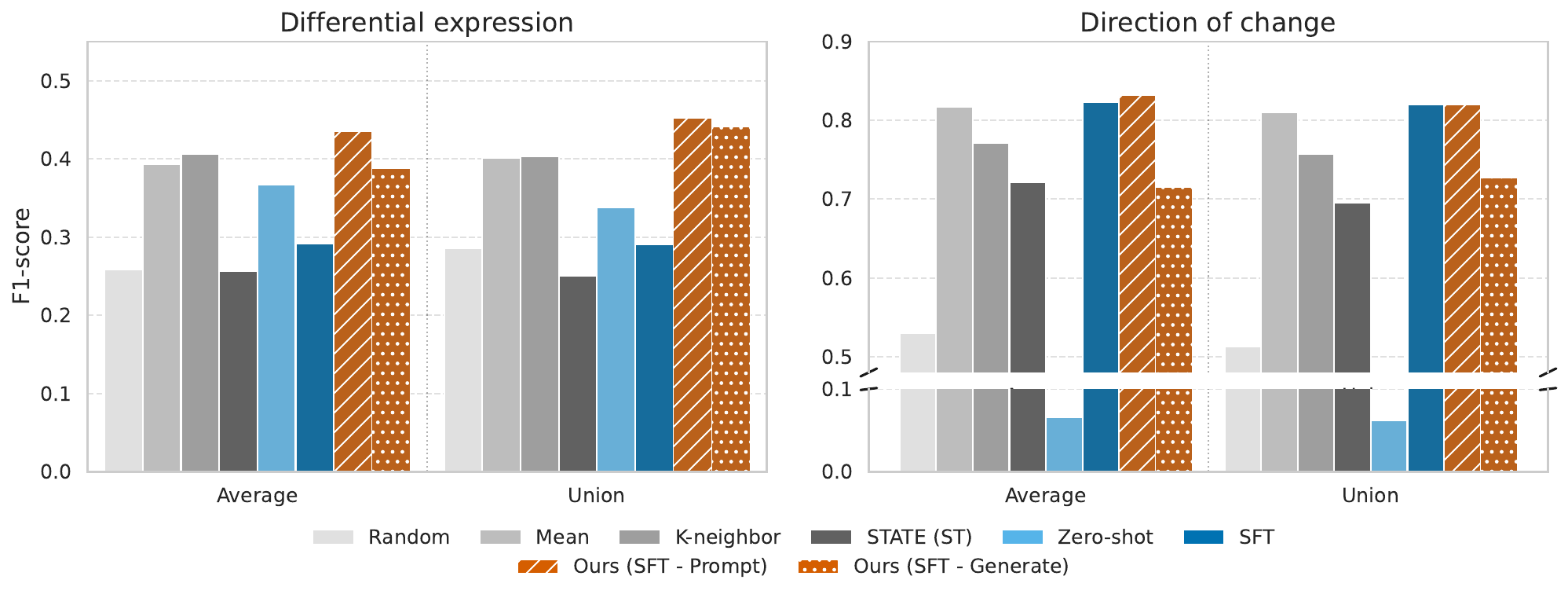}
    \vspace{-0.05in}
\caption{\textbf{TahoeQA performance.} Baselines are categorized by model type: statistical and gene foundation models are shown in shades of gray, LLM-based baselines in shades of blue, and our  model with structured explanation in \textcolor{mybrown}{brown}. Average denotes the mean F1-score  across the five individual cell-line test sets while Union denotes the performance on a test set combining all five cell lines.} \label{fig5_2:tahoeqa}
\vspace{-0.2in}
\end{figure*}

Next, we evaluate the utility of our \datasetname dataset on TahoeQA, a downstream task designed to predict transcriptional responses to chemical compounds using perturbations sourced from \mbox{Tahoe-100M}. This task is inspired by the PerturbQA benchmark~\citep{wu2024contextualizing}.

\paragraph{Dataset.} We use the same perturbation-context pairs from the Tahoe-100M dataset introduced in \cref{sec:5_1_explanation_quality}. Inspired by PerturbQA, for data labeling, we extracted the top 25 up-regulated, top 25 down-regulated, and 100 random non-regulated genes for each pair. Specifically, we perform differential expression (DE) analysis by fitting a negative binomial-based general linear model to the pseudo-bulked counts, and running a Wald's test~\citep{wald1943tests} to determine if the logFC differs significantly from 0 as implemented in DESeq2~\citep{love2014moderated}. We define differentially expressed genes following ~\citet{benjamini1995controlling} adjusted $p < 0.05$, and the top 25 DE genes are selected based on the magnitude of log$_2$ fold change. 

Based on the labeled dataset, we select five cell types following the few-shot test split of Tahoe-X1~\citep{gandhi2025tahoe}: C32, HOP62, HepG2/C3A, Hs 766T, PANC-1. Following Tahoe-X1, the data is then split by perturbation to ensure no overlap between training and test perturbations, with 1K test examples randomly selected for evaluation. {Notably, we confirm that our DE verifier does not introduce test-time label leakage: the overlap between test genes and those appearing in \texttt{regulates\_expression} actions is minimal (0.2\% for DE, 0.1\% for DOC).}

\paragraph{Task.} Following PerturbQA, the task is a two-fold binary classification: (1) differential expression and (2) direction of change. For both, perturbation and cellular context are given as an input question. The differential expression task predicts whether a perturbation causes differential expression or not, while the direction of the change task predicts whether the target gene's expression decreases or increases.

\paragraph{Baselines.} We compare our trained models against three categories of baselines: (1) simple statistical models, (2) a transcriptomic foundation model, and (3) LLMs. The statistical baselines include: (1) a random baseline, (2) a mean baseline that predicts labels based on the average gene expression response for a given compound, and (3) a $k$-nearest-neighbor baseline that performs label classification by aggregating the labels of the $k$ most similar compounds, where similarity is computed using extended-connectivity fingerprints (ECFP)~\citep{rogers2010extended}. {Notably, statistical baselines including mean and $k$-nearest neighbor baselines have proven to be a strong baseline in differential expression tasks~\citep{comparison, wenkel2025txpert}.}

For the transcriptomic foundation model, we include the STATE Transition (ST) model ~\citep{adduri2025predicting} which learns a state-transition function over gene expression from a large corpus of perturbation-response data in \mbox{Tahoe-100M}~\citep{Zhang2025.02.20.639398}. The model is trained on all available perturbations across cell types and evaluated under our few-shot setting, where test perturbations are held out in the five target cell lines and used only for evaluation. Finally, the LLM baselines include zero-shot prompting and supervised fine-tuning (SFT) without any structured explanations and use the same Qwen3 backbone as our models.

\paragraph{Training.} We fine-tune Qwen3-4B-Instruct-2507~\citep{yang2025qwen3technicalreport} to assess whether our structured reasoning serves as an effective supervision signal. We train the model using SFT in two configurations: (1) context-augmented prediction (SFT-Prompt), where the model predicts the answer label given the perturbation, cellular context, and the verified structured explanation as input; and (2) generative reasoning (SFT-Generate), where the model is trained to generate the structured explanation followed by the answer, given only the perturbation and cellular context. Notably, we focus on SFT and exclude reinforcement learning (RL) from this study, as reasoning in pure classification settings often suffers from sparse reward signals ~\citep{NEURIPS2024_ad236edc, he2025gencls++, sprague2025to} and leave RL-based optimization for future work.

\paragraph{Metrics.} We measure the performance with F1-score due to the label imbalance problem (i.e., 50 positive and 100 negative labels per perturbation-context pair) in the DE task.

\paragraph{Results.} We provide the results in \cref{fig5_2:tahoeqa} and detailed per-cell-line values in \cref{appx: exp}. Our experiments demonstrate that structured reasoning significantly and consistently enhances predictive accuracy, particularly for the DE task. Conditioning predictions on structured explanations (SFT-Prompt) yields the strongest overall performance across cell types and tasks. Additionally, our generative model (SFT-Generate), which autonomously constructs the mechanistic reasoning chain, substantially surpasses all baselines in the DE task. This performance gap between our structured approaches and standard SFT confirms that training models with explicit biological reasoning provides a more effective supervision signal than direct label prediction alone. 

Crucially, by leveraging structured reasoning as an inductive bias, our model exhibits superior generalization to novel compounds compared to baselines like STATE, which rely primarily on raw numerical representations. These findings suggest that grounding high-dimensional transcriptomic data in biological reasoning improves performance in sparse-data and out-of-distribution scenarios.
\section{Related Work}

\paragraph{Reasoning with LLMs} Reasoning of large language models (LLMs) has surprisingly enhanced the problem-solving capabilities~\citep{openai2024openaio1card,deepseekai2025deepseekr1incentivizingreasoningcapability,wei2022chain}. The reasoning models are typically trained through supervised fine-tuning or reinforcement learning, using human-curated~\citep{cobbe2021gsm8k, gao2024omni, hendrycksmath2021} reasoning as supervision signals. However, such methods rely heavily on high-quality annotated explanations, which are expensive and domain-limited.

To mitigate data scarcity, LLMs are increasingly leveraged to synthesize reasoning traces~\citep{wang-etal-2023-self-instruct, moshkov2025aimo2, guha2025openthoughtsdatarecipesreasoning}, yet ensuring factual reliability remains a major challenge. While the reasoning in mathematics and code can be validated via symbolic or programmatic evaluation~\citep{shao2024deepseekmathpushinglimitsmathematical, chen2021evaluatinglargelanguagemodels}, verification in empirical science, such as biology, is hindered by causal uncertainty and incomplete prior knowledge. As current general-purpose LLMs lack sufficient grounding in specialized scientific contexts, their self-generated explanations frequently exhibit factual inconsistency or domain hallucination, limiting their reliability.

\paragraph{Reasoning for biology} LLMs are increasingly recognized as powerful tools for scientific discovery, with applications spanning chemistry and biology~\citep{han-etal-2025-generalist, fang2024molinstructions, edwards2022molt5}, biology~\citep{zhang2024scientificlargelanguagemodels}. Yet, limited reasoning capability hinders their ability to answer complex, open-ended scientific questions. This deficiency is particularly problematic in biology, where questions must incorporate varying levels of uncertainty and reliability is hard to verify, such as predicting cellular responses to novel drugs. Recent efforts to enhance biological reasoning include RL frameworks with soft verifiers~\citep{Istrate2025.08.18.670981}, SFT on GPT-4o reasoning traces~\citep{phillips2025synthpertenhancingllmbiological}, and inference-time compression of gene-centric knowledge graphs~\citep{wu2025perturbqa}. Despite these advances, they remain largely restricted to gene-centric perturbations and often rely on unstructured, free-form natural language rationales.

Prior works face three primary limitations: (1) Lack of reliable fact discovery: existing models rely on internal parametric weights without sufficient integration of external knowledge, leading to biologically implausible hallucinations. (2) Structural ambiguity: unstructured rationales prevent the formalization of mechanistic dependencies, rendering them unsuitable for systematic verification. (3) Cross-modal insufficiency: existing strategies focus almost exclusively on gene-centric perturbations, neglecting the complexity of diverse modalities, such as drug-induced responses.

To address these challenges, we propose a paradigm shift toward structured mechanistic reasoning: treating biological reasoning as the autonomous construction of structured mechanistic graphs. By replacing free-form natural language with a verifiable assembly of discrete and biologically grounded actions, \Methodname transforms the LLM into a mechanistic architect. Our framework synthesizes diverse knowledge bases and employs automated verifiers to ensure that every node in the reasoning graph represents an evidence-aligned, falsifiable claim. This enables autonomous agents to not only predict cellular outcomes but to rigorously ground their predictions in mechanistic and autonomous reasoning traces.
\section{Conclusion}

In this work, we tackled the critical bottleneck in reasoning for scientific discovery by defining a structured, falsifiable reasoning format for virtual cells. We introduced \Methodname, a multi-agent system that generates and validates these explanations by separating knowledge retrieval from construction and employing a rigorous verifier pipeline. Our experiments demonstrate that \Methodname produces explanations with factual accuracy and logical coherence, outperforming baselines. Moreover, this high-quality dataset boosts the performance on the downstream TahoeQA gene expression prediction task. Our framework provides a scalable method for generating grounded and mechanistic reasoning, shedding light on the way to reliable and autonomous virtual cells.

\newpage
\section*{Impact statement}

The development of \Methodname addresses a critical bottleneck in the reliability of large language models for scientific tasks: the lack of interpretability and factual grounding in Large Language Models (LLMs) when applied to complex biological systems. By transitioning from unstructured text generation to structured mechanistic graphs, this work provides a framework for the systematic verification and falsification of biological hypotheses.


A primary contribution of this work is the release of \datasetname, derived from Tahoe-100M. By providing the research community with a scalable and grounded resource, we lower the barrier to entry for studying complex cellular responses to perturbations. Rather than functioning as a closed-loop solution, this dataset and framework are designed to augment domain experts, allowing them to integrate and refine mechanistic explanations within their specific research pipelines. This facilitates the development of autonomous virtual cells that can navigate cellular events with increased factual precision.

Despite the improvements in factual precision, we emphasize that the generated reasoning traces are intended for mechanistic plausibility rather than formal causal discovery or direct clinical implementation. The framework’s dependency on external knowledge bases (e.g. StarkPrimeKG~\citep{wu24stark} and Harmonizome~\citep{10.1093/nar/gkae1080}) means that inherent biases or incomplete data in these repositories may influence the validity of the reasoning output. Consequently, there is a risk of model hallucination in specialized scientific contexts where external grounding is insufficient or contradictory. 

To mitigate these risks, we employ a verifier-based filtering pipeline. While our current implementation includes only the verifiers most critical to our downstream tasks—Drug-Target Interactions and Differential Expression—this represents an extensible foundation rather than an exhaustive solution. We strongly encourage researchers to adopt and expand this verification suite to meet their specific domain requirements. Continued expansion of these capabilities across diverse biological modalities is essential to mitigate the risk of biologically implausible reasoning, which is particularly critical in sensitive biomedical research. Adherence to these rigorous validation standards ensures that autonomous virtual cells serve as reliable and ethical collaborators in the pursuit of scientific truth.

\newpage
\bibliography{reference}
\bibliographystyle{valence}

\newpage
\appendix

\appendix
\crefalias{section}{appendix}
\setcounter{figure}{0}
\renewcommand{\thefigure}{Appx.\arabic{figure}}
\setcounter{table}{0}
\renewcommand{\thetable}{Appx.\arabic{table}}

\section{Details of action primitives}\label{appx: details_action}

Here, we describe the details of action primitives. The action primitives are grouped into eight categories: system initialization,
molecular interactions, signaling \& metabolism, protein dynamics, transcription \& translation, genetic perturbations, cellular outcomes, and descriptive. We describe the role of each action below and argument in \cref{tab_appx_1:action_schema}.

\begin{table}[h]
    \centering
    \caption{\textbf{Argument schema for action spaces.} }\label{tab_appx_1:action_schema}
    \resizebox{\linewidth}{!}{
    \begin{tabular}{cccc}
    \toprule[1.25pt]
       \textbf{Category} & \textbf{Sub-category}  & \textbf{Action name} & \textbf{Arguments}  \\
       \midrule
       \multicolumn{2}{c}{system initialization} & set\_context & \texttt{set\_context}(\{cell\_type, genotype, disease, prior\_perturbation, extras\}) \\
       \midrule
        \multicolumn{2}{c}{metabolic} & converts\_substrate & \texttt{converts\_substrate}(id, enzyme, substrate, product, \{via, confidence\}) \\
        \midrule
        \multirow{7}{*}{regulation} & \multirow{2}{*}{activity modulation} & modulates\_molecule\_activity & \texttt{modulates\_molecule\_activity}(id, target, direction, \{via, confidence\}) \\
       & & modulates\_pathway\_activity & \texttt{modulates\_pathway\_activity}(id, pathway, direction, \{via, confidence\}) \\
       \cmidrule(lr){2-4}
        & \multirow{2}{*}{protein regulation} & modulates\_complex & \texttt{modulates\_complex}(id, members, complex, direction, \{stoichiometry, via, confidence\}) \\
        & & post\_translational\_modification & \texttt{post\_translational\_modification}(id, protein, mod\_type, site, direction, \{via, confidence\}) \\
        \cmidrule(lr){2-4}
        & transcriptional regulation & regulates\_expression & \texttt{regulates\_expression}(id, regulator, gene\_list, direction, \{mechanism, via, confidence\}) \\
        \cmidrule(lr){2-4}
        & translational regulation  & regulates\_translation & \texttt{regulates\_translation}(id, regulator, rna\_id, direction, \{mechanism, via, confidence\}) \\
        \cmidrule(lr){2-4}
        & epigenetic regulation & chromatin\_modification & \texttt{chromatin\_modification}(id, mark, locus, direction, \{via, confidence\}) \\
        \midrule
        \multirow{5}{*}{functional} & \multirow{2}{*}{functional perturbation} & gain\_of\_function & \texttt{gain\_of\_function}(id, variant\_id, protein, \{via, confidence\}) \\
        & & loss\_of\_function & \texttt{loss\_of\_function}(id, variant\_id, protein, \{via, confidence\})  \\
        \cmidrule(lr){2-4}
        & \multirow{3}{*}{functional association} & similar\_to & \texttt{similar\_to}(id, entity\_a, entity\_b, evidence\_type, \{confidence, via\}) \\
        & & correlates\_with & \texttt{correlates\_with}(id, entity\_a, entity\_b, evidence\_type, \{confidence\})\\
        & & participates\_in & \texttt{participates\_in}(id, entity, ontology\_id, \{evidence\_type, confidence\}) \\
        \midrule
        \multicolumn{2}{c}{\multirow{2}{*}{interaction}} & binds\_to & \texttt{binds\_to}(id, actor, target, \{affinity, unit, residues\_actor, residues\_target, via, confidence\}) \\
        & & cell\_cell\_interaction & \texttt{cell\_cell\_interaction}(id, sender, receiver, ligand, receptor, outcome, \{via, confidence\}) \\
        \midrule
        \multicolumn{2}{c}{\multirow{2}{*}{phenotype}} & induces\_phenotype & \texttt{induces\_phenotype}(id, source, phenotype, \{via, confidence, from\_state, to\_state\}) \\
       & & alleviates\_phenotype & \texttt{alleviates\_phenotype}(id, actor, phenotype, \{via, confidence, from\_state, to\_state\}) \\
       \midrule
        \multicolumn{2}{c}{\multirow{2}{*}{proteostasis}} & localizes\_to & \texttt{localizes\_to}(id, entity, from\_loc, to\_loc, \{mechanism, via, confidence\}) \\
       & & degrades\_or\_stabilizes & \texttt{degrades\_or\_stabilizes}(id, regulator, target, direction, \{via, confidence\}) \\
     \bottomrule[1.25pt] 
    \end{tabular}
    }
\end{table}

\subsection{System initialization}
\begin{itemize}
    \item \texttt{set\_context}: Defines the biological background before the new perturbation is applied including the cell model, genotype, disease, and prior treatments.
\end{itemize}

\subsection{Metabolic}

\begin{itemize}
    \item \texttt{converts\_substrate}: Enzymatic conversion of one chemical entity into another (metabolic reaction or proteolytic processing).
\end{itemize}

\subsection{Regulation}

\paragraph{Activity modulation}
\begin{itemize}
     \item \texttt{modulates\_molecule\_activity}: Describes whether increases or decreases the catalytic / signalling activity of a single protein, enzyme, transporter, TF or RNA.
    \item \texttt{modulates\_pathway\_activity}: Increases or decreases the activity of a named pathway or biological process (e.g. MAPK, autophagy, EMT).
\end{itemize}

\paragraph{Protein regulation}

\begin{itemize}
    \item \texttt{modulates\_complex}: Promotes assembly or causes disassembly of a multi-subunit complex; can include stoichiometry changes.
    \item \texttt{post\_translational\_
modification}: Adds or removes a specific PTM or proteolytic cleavage on a protein site (phospho, ubiquitin, acetyl, etc.)
\end{itemize}

\paragraph{Transcriptional regulation}
\begin{itemize}
    \item \texttt{regulates\_expression}: Alters steady-state mRNA level or isoform ratio of one or more genes (bulk, single-cell or signature). Use `mechanism` to tag cases like alternative splicing.
\end{itemize}

\paragraph{Translation regulation}
\begin{itemize}
    \item \texttt{regulates\_translation}: Post-transcriptional control at the ribosome level (e.g. eIF inhibition, uORF usage, IRES activation).
\end{itemize}

\paragraph{Epigenetic regulation}
\begin{itemize}
    \item \texttt{chromatin\_modification}: Adds or removes histone/DNA marks at a defined genomic locus, affecting chromatin accessibility.
\end{itemize}

\subsection{Functional}

\paragraph{Functional perturbation}
\begin{itemize}
    \item \texttt{gain\_of\_function}: Genetic variant or edit that increases the normal activity of the specified protein.
    \item \texttt{loss\_of\_function}: Genetic variant, knock-down or knock-out that reduces or abolishes activity of the specified protein.
\end{itemize}

\paragraph{Functional association}
\begin{itemize}
    \item \texttt{similar\_to}: Non-causal functional similarity (transcriptomic, phenotypic, structural) between two entities.
\item \texttt{correlates\_with}: Statistical association without established causality (GWAS hit, literature co-mention, co-expression).
\item \texttt{participates\_in}: Links an entity to a GO term, pathway or compartment (background annotation).
\end{itemize}

\subsection{Interaction}

\begin{itemize}
    \item \texttt{binds\_to}: Describes the direct physical binding of two biomolecules such as drug-target, protein-protein, and ligand-receptor.
    \item \texttt{cell\_cell\_interaction}: Represents the ligand-receptor signaling from one cell type to another plus the immediate downstream outcome.
\end{itemize}

\subsection{Phenotype}
\begin{itemize}
    \item \texttt{induces\_phenotype}: Creates or worsens a measurable phenotype or cell-state transition. Optional 'from\_state' / 'to\_state' capture events like EMT or senescence.
    \item \texttt{alleviates\_phenotype}: Reverts or mitigates an abnormal phenotype back toward normal (rescue). Same optional state fields as above.
\end{itemize}

\subsection{Proteostasis}

\begin{itemize}    
\item \texttt{localizes\_to}: Moves a molecule between compartments. Use 'to="extracellular"' for secretion or 'from="extracellular"' for uptake; add 'mechanism' (e.g. exocytosis, transporter).
\item \texttt{degrades\_or\_stabilizes}: Changes protein abundance by altering half-life (ubiquitin-proteasome degradation, PROTAC, chaperone rescue).
\end{itemize}

\newpage
\section{Examples}\label{appx: example}

\subsection{Retrieved information}

\paragraph{StarkPrimeKG.} This is the example of retrieved information from StarkPrimeKG. Note that due to the limited space, we display only partial information of the full retrieved context.

  \begin{tcolorbox}[
    colback=gray!5!white, 
    colframe=gray!60!teal,
    title={
      \parbox[t]{\dimexpr\linewidth-4mm\relax}{%
        \ttfamily
        Retrieved information from StarkPrimeKG for $(p,c)=$(Binimetinib, C32)
      }
    },
    breakable
  ]
\#\# KNOWLEDGE GRAPH INFORMATION
\vspace{\baselineskip}

- name: MAP2K1 

    - type: gene/protein 
    
    - source: NCBI - details:   
    
    - query: MAP2K1   
    
    - alias (other gene names): [`CFC3', `MAPKK1', `MEK1', `MEL', `MKK1', `PRKMK1']   
    
    - genomic\_pos (genomic position): {`chr': '15', `end': 66491656, `ensemblgene': `ENSG00000169032', `start': 66386837, `strand': 1}
    
    - name (gene name): mitogen-activated protein kinase kinase 1  
    
    - summary (protein summary text): The protein encoded by this gene is a member of the dual specificity protein kinase family, which acts as a mitogen-activated protein (MAP) kinase kinase. MAP kinases, also known as extracellular signal-regulated kinases (ERKs), act as an integration point for multiple biochemical signals. This protein kinase lies upstream of MAP kinases and stimulates the enzymatic activity of MAP kinases upon wide variety of extra- and intracellular signals. As an essential component of MAP kinase signal transduction pathway, this kinase is involved in many cellular processes such as proliferation, differentiation, transcription regulation and development. [provided by RefSeq, Jul 2008].
    
    - relations:   
    
        ppi: {gene/protein: (EGR1, MYC, GSK3B, PIK3R1, AURKA, CTCF, CHEK1, ETS1, E2F4, UBE2I, UBC, MAPK14, ARRB2, CDKN2A, \dots),} 
        
        target: {drug: (Selumetinib, Cobimetinib, Bosutinib, Trametinib, K-252a, 5-Bromo-N-[(2S)-2,3-dihydroxypropoxy]-3,4-difluoro-2-[(2-fluoro-4-iodophenyl)amino]benzamide, (5S)-4,5-difluoro-6-[(2-fluoro-4-iodophenyl)imino]-N-(2-hydroxyethoxy)cyclohexa-1,3-diene-1-carboxamide, 2-[(2-chloro-4-iodophenyl)amino]-N-{[(2R)-2,3-dihydroxypropyl]oxy}-3,4-difluorobenzamide, \dots)}   
        
        associated\_with: {disease: (Noonan syndrome, Prader-Willi syndrome, anxiety disorder, cardiofaciocutaneous syndrome, melanoma, cutaneous malignant, susceptibility to, Costello syndrome, LEOPARD syndrome, Noonan syndrome-like disorder with loose anagen hair, familial prostate carcinoma, prostate cancer, hereditary, lung cancer, ovarian cancer, hereditary breast ovarian cancer syndrome, Noonan syndrome with multiple lentigines, melanoma, non-small cell lung carcinoma (disease), lung adenocarcinoma, ovarian neoplasm, hairy cell leukemia, transient ischemic attack (disease), yolk sac tumor, colorectal adenocarcinoma, gonadoblastoma, ovarian mucinous adenocarcinoma, ovarian adenocarcinoma, \dots,}   
        
        interacts\_with: {cellular\_component: (endoplasmic reticulum, late endosome, microtubule organizing center, nucleus, Golgi apparatus, plasma membrane, early endosome, cytosol, mitochondrion, focal adhesion),molecular\_function: (protein serine/threonine kinase activity, protein binding, protein kinase activity, transmembrane receptor protein tyrosine kinase activity, protein serine/threonine/tyrosine kinase activity, MAP kinase kinase activity, protein serine/threonine kinase activator activity, protein N-terminus binding, protein threonine kinase activity, protein serine kinase activity, protein C-terminus binding, scaffold protein binding, ATP binding, MAP-kinase scaffold activity),pathway: (MAPK3 (ERK1) activation, \dots ,} \\

- name: Binimetinib 

- type: drug 

- source: DrugBank 

- details:   

    - description: Binimetinib, also known as \_Mektovi\_, is a potent is a potent and selective oral mitogen-activated protein kinase 1/2  (MEK 1/2) inhibitor.   
    
    - half\_life: The mean (CV\%) terminal half-life (t1/2) of binimetinib is 3.5 hours (28.5\%).   
    
    - protein\_binding: Binimetinib is 97\% bound to human plasma proteins and the blood-to-plasma ratio is 0.72 
    
    - pharmacodynamics: Binimetinib is a MEK inhibitor. MEK is an enzyme that regulates the biosynthesis of the inflammatory cytokines TNF, IL-6 and IL-1. MEK inhibitors interfere with these biosynthetic processes.  It is a chemotherapeutic agent that has anti-tumor activity.   
    
    - state: Binimetinib is a solid.   
        - atc\_1: Binimetinib is anatomically related to antineoplastic and immunomodulating agents.   
        - atc\_2: Binimetinib is in the therapeutic group of antineoplastic agents.   
        - atc\_3: Binimetinib is pharmacologically related to other antineoplastic agents. 
        - atc\_4: The chemical and functional group of  is protein kinase inhibitors.   
        
    - category: Binimetinib is part of Antineoplastic Agents ; Antineoplastic and Immunomodulating Agents ; Cytochrome P-450 CYP1A2 Substrates ; Cytochrome P-450 CYP1A2 Substrates with a Narrow Therapeutic Index ; Cytochrome P-450 CYP2C19 Substrates ; Cytochrome P-450 CYP2C19 Substrates with a Narrow Therapeutic Index ; Cytochrome P-450 Substrates ; Heterocyclic Compounds, Fused-Ring ; MAP Kinase Kinase 1, antagonists \& inhibitors ; MAP Kinase Kinase 2, antagonists \& inhibitors ; Narrow Therapeutic Index Drugs ; P-glycoprotein substrates with a Narrow Therapeutic Index ; Protein Kinase Inhibitors ; UGT1A1 Substrates ; UGT1A1 Substrates with a Narrow Therapeutic Index.  
    
    - group: Binimetinib is approved and investigational. 
    
    - relations:   
    
    enzyme: {gene/protein: (CYP1A2, CYP2C19, UGT1A1),}   
    
    target: {gene/protein: (IL1B, IL6, TNF, MAP3K1, MAP2K2),}   
    
    indication: {disease: (metastatic melanoma),}   
    
    synergistic\_interaction: {drug: (Prednisolone phosphate, Levothyroxine, Diclofenac, Genistein, Resveratrol, Nelfinavir, Nevirapine, Phenytoin, Conjugated estrogens, Morphine, Desogestrel, Valproic acid, Acetaminophen, Amitriptyline, Indomethacin, Olanzapine, Rosiglitazone, Meperidine, Imipramine, Nabumetone, Fluoxetine, Duloxetine, Chlorpromazine, Zidovudine, Ritonavir, Erlotinib, Ciprofloxacin, \dots,}

    \dots
  \end{tcolorbox}

\paragraph{Harmonizome.} This is the example of retrieved information from Harmonizome. Note that due to the limited space, we display only partial information of the full retrieved context.

  \begin{tcolorbox}[
    colback=gray!5!white, 
    colframe=gray!60!teal,
    title={
      \parbox[t]{\dimexpr\linewidth-4mm\relax}{%
        \ttfamily
        Retrieved information from Harmonizome for $(p,c)=$(Binimetinib, C32)
      }
    },
    breakable
  ]
\#\# GENE INFORMATION

\vspace{\baselineskip}
\#\#\# TARGET GENE INFORMATION

\vspace{\baselineskip}

The target gene of the perturbation is MAP2K1. The gene name is mitogen-activated protein kinase kinase 1 and the gene symbol is MAP2K1.The protein encoded by this gene is a member of the dual specificity protein kinase family, which acts as a mitogen-activated protein (MAP) kinase kinase. MAP kinases, also known as extracellular signal-regulated kinases (ERKs), act as an integration point for multiple biochemical signals. This protein kinase lies upstream of MAP kinases and stimulates the enzymatic activity of MAP kinases upon wide variety of extra- and intracellular signals. As an essential component of MAP kinase signal transduction pathway, this kinase is involved in many cellular processes such as proliferation, differentiation, transcription regulation and development. [provided by RefSeq, Jul 2008]The gene synonyms are MKK1, MEK1, CFC3, PRKMK1, MAPKK1.The gene is related to the proteins MP2K1\_HUMAN.

\vspace{\baselineskip}
\#\#\# CELL TYPE INFORMATION

\vspace{\baselineskip}

The cell type of the context is C32. The related genes are GANAB, CNOT11, HNRNPR, MAP2K1, CASP3, HABP2, NUAK2, KRTCAP2, IL6, BBS9, LARGE1, AKT1, SGMS1, TYR, RARS1, CERS2, NPL, RAN, PXDN, PRDM16.  The gene information is as follows:

The gene name is glucosidase, alpha; neutral AB and the gene symbol is GANAB.This gene encodes the alpha subunit of glucosidase II and a member of the glycosyl hydrolase 31 family of proteins. The heterodimeric enzyme glucosidase II plays a role in protein folding and quality control by cleaving glucose residues from immature glycoproteins in the endoplasmic reticulum. Expression of the encoded protein is elevated in lung tumor tissue and in response to UV irradiation. Mutations in this gene cause autosomal-dominant polycystic kidney and liver disease. [provided by RefSeq, Jul 2016]The gene synonyms are GLUII, G2AN, GIIA.The gene is related to the proteins GANAB\_HUMAN.

The gene name is CCR4-NOT transcription complex, subunit 11 and the gene symbol is CNOT11.Predicted to be involved in nuclear-transcribed mRNA poly(A) tail shortening. Part of CCR4-NOT complex. [provided by Alliance of Genome Resources, Mar 2025]The gene synonyms are C40, C2ORF29.The gene is related to the proteins CNO11\_HUMAN.

The gene name is heterogeneous nuclear ribonucleoprotein R and the gene symbol is HNRNPR.This gene encodes an RNA-binding protein that is a member of the spliceosome C complex, which functions in pre-mRNA processing and transport. The encoded protein also promotes transcription at the c-fos gene. Alternative splicing results in multiple transcript variants. There are pseudogenes for this gene on chromosomes 4, 11, and 10. [provided by RefSeq, Jul 2014] The gene synonyms are HNRPR, HNRNP-R. The gene is related to the proteins HNRPR\_HUMAN.

The gene name is mitogen-activated protein kinase kinase 1 and the gene symbol is MAP2K1. The protein encoded by this gene is a member of the dual specificity protein kinase family, which acts as a mitogen-activated protein (MAP) kinase kinase. MAP kinases, also known as extracellular signal-regulated kinases (ERKs), act as an integration point for multiple biochemical signals. This protein kinase lies upstream of MAP kinases and stimulates the enzymatic activity of MAP kinases upon wide variety of extra- and intracellular signals. As an essential component of MAP kinase signal transduction pathway, this kinase is involved in many cellular processes such as proliferation, differentiation, transcription regulation and development. [provided by RefSeq, Jul 2008] The gene synonyms are MAPKK1, PRKMK1, MKK1, CFC3, MEK1. The gene is related to the proteins MP2K1\_HUMAN.

\dots

  \end{tcolorbox}

\paragraph{PubMed.} This is the example of retrieved information from PubMed. Note that due to the limited space, we display only partial information of the full retrieved context.

  \begin{tcolorbox}[
    colback=gray!5!white, 
    colframe=gray!60!teal,
    title={
      \parbox[t]{\dimexpr\linewidth-4mm\relax}{\%
        \ttfamily
        Retrieved information from PubMed for $(p,c)=$(Binimetinib, C32)
      }
    },
    breakable]
\#\# RELATED PAPER LIST
\vspace{\baselineskip}

\{
    ``title": "Binimetinib, a novel MEK1/2 inhibitor, exerts anti-leukemic effects under inactive status of PI3Kinase/Akt pathway.",
    ``abstract": "A MEK1/2 inhibitor, binimetinib is promising as a therapeutic agent for malignant melanoma with N-RAS mutation. We examined in vitro effects of binimetinib on 10 human myeloid/lymphoid leukemia cell lines, and found that three of five cell lines with N-RAS mutation and one of five without N-RAS mutation were responsive to treatment with binimetinib. Binimetinib inhibited cell growth mainly by inducing G"
\}

\{
    ``title": "Severe Drug-Induced Liver Injury from Combination Encorafenib/Binimetinib.",
    ``abstract": "Encorafenib/binimetinib is a new combination BRAF/MEK inhibitor used in the treatment of advanced or metastatic BRAFV600-mutant melanoma. Though generally tolerated well, mild to moderate aminotransferase elevations are common. However, significant liver injury has not been demonstrated in the literature. Here, we report the first case of severe hepatic injury associated with encorafenib/binimetinib in a 58-year-old gentleman requiring admission and extensive workup. He was successfully treated by withdrawing the combination therapy, and liver function returned to normal range."
\}

\{
    ``title": "Phase Ib Study of Combination Therapy with MEK Inhibitor Binimetinib and Phosphatidylinositol 3-Kinase Inhibitor Buparlisib in Patients with Advanced Solid Tumors with RAS/RAF Alterations.",
    ``abstract": "This multicenter, open-label, phase Ib study investigated the safety and efficacy of binimetinib (MEK inhibitor) in combination with buparlisib (phosphatidylinositol 3-kinase [PI3K] inhibitor) in patients with advanced solid tumors with RAS/RAF alterations."
\}

\{
    ``title": "Encorafenib and binimetinib for the treatment of BRAF-mutated metastatic melanoma in the setting of combined hepatic and renal impairment.",
    ``abstract": "Inhibitors of BRAF, a gene coding a protein called B-raf, with or without inhibitors of MEK (MAPK/extracellular signal-regulated kinase) are often used as palliative treatment in BRAF-mutated metastatic melanoma. Recent data show improved progression-free survival with encorafenib with binimetinib, a newer BRAF/MEK inhibitor combination, compared with older agents, but there have been no reports of this treatment in the setting of renal and liver failure. We report a patient with disease-induced transaminitis and renal failure requiring dialysis who was successfully treated with encorafenib and binimetinib. His transaminitis improved and he was able to stop dialysis without any significant adverse effects during treatment, suggesting encorafenib with binimetinib may be used safely and effectively even in patients with end organ damage."
\}

\{
    ``title": "MEK Inhibitors in the Treatment of Metastatic Melanoma and Solid Tumors.",
    ``abstract": "The mitogen-activated protein kinase (MAPK) cascade is an intracellular signaling pathway involved in the regulation of cellular proliferation and the survival of tumor cells. Several different mutations, involving BRAF or NRAS, exert an oncogenic effect by activating the MAPK pathway, resulting in an increase in cellular proliferation. These mutations have become targets for new therapeutic strategies in melanoma and other cancers. Selective MEK inhibitors have the ability to inhibit growth and induce cell death in BRAF- and NRAS-mutant melanoma cell lines. MEK inhibitor therapy in combination with a BRAF inhibitor is more effective and less toxic than treatment with a BRAF inhibitor alone, and has become the standard of care for patients with BRAF-mutated melanoma. Trametinib was the first MEK inhibitor approved for the treatment of BRAF-mutated metastatic melanoma not previously treated with BRAF inhibitors, and is also approved in combination with the BRAF inhibitor dabrafenib. Furthermore, cobimetinib is another MEK inhibitor approved for the treatment of BRAF-mutated metastatic melanoma in combination with a BRAF inhibitor, vemurafenib. The MEK inhibitor binimetinib in combination with the BRAF inhibitor encorafenib is in clinical development. The addition of an anti-PD-1/PD-L1 agent, such as pembrolizumab, durvalumab or atezolizumab, to combined BRAF and MEK inhibition has shown considerable promise, with several trials ongoing in metastatic melanoma. Binimetinib has also shown efficacy in NRAS-mutated melanoma patients. Future possibilities for MEK inhibitors in advanced melanoma, as well as other solid tumors, include their use in combination with other targeted therapies (e.g. anti-CDK4/6 inhibitors) and/or various immune-modulating antibodies."
\}
...

  \end{tcolorbox}

\paragraph{Wikipedia.} This is the example of retrieved information from Wikipedia. Note that due to the limited space, we display only partial information of the full retrieved context.

  \begin{tcolorbox}[
    colback=gray!5!white, 
    colframe=gray!60!teal,
    title={
      \parbox[t]{\dimexpr\linewidth-4mm\relax}{\%
        \ttfamily
        Retrieved information from Wikipedia for $(p,c)=$(Binimetinib, C32)
      }
    },
    breakable
]

\#\# WIKIPEDIA INFORMATION
\vspace{\baselineskip}

\#\# MAP2K2

Dual specificity mitogen-activated protein kinase kinase 2 is an enzyme that in humans is encoded by the MAP2K2 gene. It is more commonly known as MEK2, but has many alternative names including CFC4, MKK2, MAPKK2 and PRKMK2.

== Function ==

The protein encoded by this gene is a dual specificity protein kinase that belongs to the MAP kinase kinase family. This kinase is known to play a critical role in mitogen growth factor signal transduction. It phosphorylates and thus activates MAPK1/ERK2 and MAPK3/ERK1.
The activation of this kinase itself is dependent on the Ser/Thr phosphorylation by MAP kinase kinase kinases.
The inhibition or degradation of this kinase is found to be involved in the pathogenesis of Yersinia and anthrax.

== Interactions ==

MAP2K2 has been shown to interact with MAPK3 and ARAF.

== External links ==

GeneReviews/NCBI/NIH/UW entry on Cardiofaciocutaneous Syndrome
Overview of all the structural information available in the PDB for UniProt: P36507 (Dual specificity mitogen-activated protein kinase kinase 2) at the PDBe-KB.

\vspace{\baselineskip}

\#\# Binimetinib

Binimetinib, sold under the brand name Mektovi, is an anti-cancer medication used to treat various cancers. Binimetinib is a selective inhibitor of MEK, a central kinase in the tumor-promoting MAPK pathway. Inappropriate activation of the pathway has been shown to occur in many cancers. In June 2018 it was approved by the FDA in combination with encorafenib for the treatment of patients with unresectable or metastatic BRAF V600E or V600K mutation-positive melanoma. In October 2023, it was approved by the FDA for treatment of NSCLC with a BRAF V600E mutation in combination with encorafenib. It was developed by Array Biopharma.

== Mechanism of action ==

Binimetinib is an orally available inhibitor of mitogen-activated protein kinase kinase (MEK), or more specifically, a MAP2K inhibitor.  MEK is part of the RAS pathway, which is involved in cell proliferation and survival. MEK is upregulated in many forms of cancer. Binimetinib, uncompetitive with ATP, binds to and inhibits the activity of MEK1/2 kinase, which has been shown to regulate several key cellular activities including proliferation, survival, and angiogenesis. MEK1/2 are dual-specificity threonine/tyrosine kinases that play key roles in the activation of the RAS/RAF/MEK/ERK pathway and are often upregulated in a variety of tumor cell types. Inhibition of MEK1/2 prevents the activation of MEK1/2 dependent effector proteins and transcription factors, which may result in the inhibition of growth factor-mediated cell signaling. As demonstrated in preclinical studies, this may eventually lead to an inhibition of tumor cell proliferation and an inhibition in production of various inflammatory cytokines including interleukin-1, -6 and tumor necrosis factor.

== Development ==

In 2015, it was in phase III clinical trials for ovarian cancer, BRAF mutant melanoma, and NRAS Q61 mutant melanoma.
In December 2015, the company announced that the mutant-NRAS melanoma trial was successful. In the trial, those receiving binimetinib had a median progression-free survival of 2.8 months versus 1.5 months for those on the standard dacarbazine treatment. NDA submitted Jun 2016, and the FDA should decide by 30 June 2017.
In April 2016, it was reported that the phase III trial for low-grade ovarian cancer was terminated due to lack of efficacy.
In 2017, the FDA informed Array Biopharma that the phase III trial data was not sufficient and the New Drug Application was withdrawn.
In June 2018, it was approved for the treatment of certain melanomas by the U.S. Food and Drug Administration (FDA) in combination with encorafenib. The FDA approved binimetinib based primarily on evidence from one clinical trial (NCT01909453) of 383 patients with BRAF V600 mutation-positive melanoma that was advanced or could not be removed by surgery. The trial was conducted at 162 sites in Europe, North America, and various countries around the world.
In October 2023, the US Food and Drug Administration approved encorafenib with binimetinib for adults with metastatic non-small cell lung cancer (NSCLC) with a BRAF V600E mutation, as detected by an FDA-approved test.

\vspace{\baselineskip}

\#\# MAP2K1

Dual specificity mitogen-activated protein kinase kinase 1 is an enzyme that in humans is encoded by the MAP2K1 gene.

== Function ==

The protein encoded by this gene is a member of the dual-specificity protein kinase family that acts as a mitogen-activated protein (MAP) kinase kinase. MAP kinases, also known as extracellular signal-regulated kinases (ERKs), act as an integration point for multiple biochemical signals. This protein kinase lies upstream of MAP kinases and stimulates the enzymatic activity of MAP kinases upon activation by a wide variety of extra- and intracellular signals. As an essential component of the MAP kinase signal transduction pathway, this kinase is involved in many cellular processes such as proliferation, differentiation, transcription regulation and development. MAP2K1 is altered in 1.05\% of all human cancers.

== Meiosis ==

The genomes of diploid organisms in natural populations are highly polymorphic for insertions and deletions.  During meiosis double-strand breaks (DSBs) that form within such polymorphic regions must be repaired by inter-sister chromatid exchange, rather than by inter-homolog exchange.  Molecular-level studies of recombination during budding yeast meiosis have shown that recombination events initiated by DSBs in regions that lack corresponding sequences in the homolog are efficiently repaired by inter-sister chromatid recombination.  This recombination occurs with the same timing as inter-homolog recombination, but with reduced (2- to 3-fold) yields of joint molecules.
MAP2K1 is also known as MEK1 (see Mitogen-activated protein kinase kinase).  MEK1 is a meiotic chromosome-axis-associated kinase that is thought to slow down, but not entirely block, sister chromatid recombination. Loss of MEK1 allows inter-sister DSB repair and also inter-sister Holliday junction intermediates to increase. Despite the normal activity of MEK1 in reducing inter-sister chromatid recombination, such recombination still occurs frequently during normal budding yeast meiosis (although not as frequently as during mitosis), and up to one-third of all recombination events are between sister chromatids.

== Interactions ==

MAP2K1 has been shown to interact with C-Raf, Phosphatidylethanolamine binding protein 1, MAP2K1IP1, GRB10, MAPK3, MAPK8IP3, MAPK1 MP1, and MAP3K1.

  \end{tcolorbox}

\subsection{Generated report}

Here, we provide an example generated report.

  \begin{tcolorbox}[
    colback=gray!5!white, 
    colframe=gray!60!teal,
    title={
      \parbox[t]{\dimexpr\linewidth-4mm\relax}{%
        \ttfamily
        Retrieved information from Harmonizome for $(p,c)=$(Binimetinib, C32)
      }
    },
    breakable
  ]
 \# Comprehensive Mechanistic Report: Binimetinib Treatment in C32 Melanoma Cells
 \vspace{\baselineskip}
 
 \#\# 1. Perturbation Description
 
 \textbf{Binimetinib} is a potent and selective oral MEK1/2 inhibitor (MAP2K1/MAP2K2 inhibitor) with the SMILES structure CN1C=NC2=C1C=C(C(=C2F)NC3=C(C=C(C=C3)Br)F)C(=O)NOCCO. It functions as an \textbf{ATP-uncompetitive allosteric inhibitor} that binds to a unique allosteric site on MEK1/2, distinct from the ATP-binding pocket. This mechanism allows sustained inhibition even under high ATP concentrations typical in cancer cells.
 
 \textbf{Key pharmacological properties:}
 \begin{itemize}
   \item \textbf{IC50 values}: $\sim$12 nM for MEK1, $\sim$4.6 nM for MEK2
   \item \textbf{Mechanism}: Non-competitive with ATP, binds to inactive MEK1/2 conformation
   \item \textbf{Selectivity}: >100-fold selectivity for MEK1/2 over other kinases
   \item \textbf{Half-life}: 3.5 hours with 97\% plasma protein binding
 \end{itemize}
\vspace{\baselineskip}

 \#\# 2. Cellular Context: C32 Melanoma Cell Line
 
 The \textbf{C32 melanoma cell line} harbors critical oncogenic mutations that create a hyperactive MAPK pathway:
 
 \textbf{Driver mutations present:}
 \begin{itemize}
   \item \textbf{BRAF V600E (GoF)}: Constitutively active BRAF kinase ($\sim$500-fold increased activity)
   \item \textbf{CDKN2A loss (tumor suppressor)}: Loss of p16INK4A and p14ARF tumor suppressors
   \item \textbf{PTEN loss (LoF)}: Hyperactive PI3K/AKT survival pathway
   \item \textbf{TERT activation (GoF)}: Enhanced telomerase activity promoting immortalization
 \end{itemize}
 
 This mutational landscape creates \textbf{dual pathway hyperactivation}: MAPK (proliferation) and PI3K/AKT (survival), with compromised cell cycle control and DNA damage response.
 \vspace{\baselineskip}
 
 \#\# 3. Step-by-Step Mechanistic Analysis

 \#\#\# Step 1: Direct Molecular Target Engagement
 \textbf{Relationship: CAUSAL (direct inhibition)}
 
 Binimetinib directly binds to the allosteric site of both MAP2K1 (MEK1) and MAP2K2 (MEK2) in their inactive conformations. The drug:
 \begin{itemize}
   \item Locks MEK1/2 in a catalytically inactive state
   \item Prevents conformational changes required for kinase activation
   \item Blocks MEK1/2 phosphorylation by upstream BRAF V600E
   \item Results in >95\% inhibition of MEK1/2 kinase activity at therapeutic concentrations
 \end{itemize}
 
 \#\#\# Step 2: Disruption of MAPK Signaling Cascade
 \textbf{Relationship: CAUSAL (direct pathway blockade)}
 
 MEK1/2 inhibition immediately disrupts the hyperactive MAPK pathway:
 
 \textbf{BRAF V600E $\rightarrow$ [BLOCKED] MEK1/2 $\rightarrow$ [BLOCKED] ERK1/2}
 
 \begin{itemize}
   \item \textbf{ERK1/2 phosphorylation}: Dramatic reduction (>90\%) in phospho-ERK1/2 (Thr202/Tyr204)
   \item \textbf{ERK nuclear translocation}: Prevented, blocking transcriptional programs
   \item \textbf{Downstream kinase activity}: RSK1/2, MSK1/2 activation blocked
   \item \textbf{Timeline}: ERK dephosphorylation occurs within 1--2 hours of treatment
 \end{itemize}
 
 \#\#\# Step 3: Transcriptional Reprogramming
 \textbf{Relationship: CAUSAL (loss of ERK-dependent transcription)}
 
 ERK1/2 inhibition leads to immediate transcriptional changes:
 
 \textbf{Early response genes (1--4 hours):}
 \begin{itemize}
   \item \textbf{c-MYC downregulation}: Loss of ERK-mediated MYC stabilization and transcription
   \item \textbf{Cyclin D1 (CCND1) downregulation}: Reduced ERK/RSK-mediated transcriptional activation
   \item \textbf{c-FOS/c-JUN reduction}: Decreased AP-1 transcriptional activity
   \item \textbf{EGR1 downregulation}: Loss of immediate early response
 \end{itemize}
 
 \textbf{Cell cycle genes (4--12 hours):}
 \begin{itemize}
   \item \textbf{CDK4/6 downregulation}: Reduced transcription and protein levels
   \item \textbf{E2F target genes}: Decreased expression due to Rb hypophosphorylation
   \item \textbf{S-phase genes}: MCM2-7, PCNA, DNA polymerase subunits reduced
 \end{itemize}
 
 \#\#\# Step 4: Cell Cycle Arrest Mechanisms
 \textbf{Relationship: CAUSAL (loss of proliferative signals + intact p53-independent mechanisms)}
 
 Despite CDKN2A loss, multiple mechanisms enforce cell cycle arrest:
 
 \textbf{G1/S checkpoint activation:}
 \begin{itemize}
   \item \textbf{Rb protein hypophosphorylation}: Due to reduced CDK4/6 activity
   \item \textbf{E2F1 sequestration}: Hypophosphorylated Rb binds and inactivates E2F1
   \item \textbf{S-phase entry blocked}: Reduced origin licensing and DNA replication machinery
 \end{itemize}
 
 \textbf{Additional arrest mechanisms:}
 \begin{itemize}
   \item \textbf{p21 upregulation}: ERK-independent, stress-induced CDKN1A transcription
   \item \textbf{p27 stabilization}: Reduced ERK-mediated p27 degradation signals
   \item \textbf{ATM/ATR activation}: Replication stress responses despite compromised DNA damage checkpoints
 \end{itemize}
 
 \#\#\# Step 5: Metabolic Reprogramming
 \textbf{Relationship: CAUSAL (loss of ERK-mediated metabolic control)}
 
 MEK inhibition disrupts cancer cell metabolism:
 
 \textbf{Glycolytic pathway:}
 \begin{itemize}
   \item \textbf{Reduced glucose uptake}: Decreased GLUT1 expression and hexokinase activity
   \item \textbf{Glycolytic enzyme downregulation}: PKM2, LDHA expression reduced
   \item \textbf{ATP production decreased}: Shift from aerobic glycolysis toward oxidative phosphorylation
 \end{itemize}
 
 \textbf{Anabolic pathways:}
 \begin{itemize}
   \item \textbf{Protein synthesis reduction}: Decreased mTOR signaling and ribosome biogenesis
   \item \textbf{Lipid synthesis impairment}: Reduced SREBP1 activity and fatty acid synthesis
   \item \textbf{Nucleotide synthesis}: Decreased UMPS, CTPS1 expression
 \end{itemize}
 
 \#\#\# Step 6: Compensatory Pathway Activation
 \textbf{Relationship: CAUSAL (feedback mechanisms and bypass signaling)}
 
 Cells attempt to compensate for MAPK blockade:
 
 \textbf{RTK upregulation:}
 \begin{itemize}
   \item \textbf{EGFR/PDGFR overexpression}: Transcriptional and post-transcriptional upregulation
   \item \textbf{Insulin/IGF-1R activation}: Enhanced survival signaling
   \item \textbf{Timeline}: Begins 12--24 hours post-treatment
 \end{itemize}
 
 \textbf{PI3K/AKT hyperactivation:}
 \begin{itemize}
   \item \textbf{Enhanced AKT phosphorylation}: Loss of ERK-mediated negative feedback on RTKs
   \item \textbf{mTOR complex reactivation}: Compensatory survival signaling
   \item \textbf{FOXO inactivation}: Maintained via AKT despite ERK loss
 \end{itemize}
 
 \#\#\# Step 7: Apoptotic Priming and Cell Death
 \textbf{Relationship: CAUSAL (inadequate survival signaling + metabolic stress)}
 
 Extended MEK inhibition triggers apoptotic cascades:
 
 \textbf{Mitochondrial dysfunction:}
 \begin{itemize}
   \item \textbf{BCL-2 family imbalance}: Increased BIM, PUMA; decreased MCL-1, BCL-2
   \item \textbf{Cytochrome c release}: Mitochondrial outer membrane permeabilization
   \item \textbf{ATP depletion}: Metabolic crisis from glycolytic shutdown
 \end{itemize}
 
 \textbf{Caspase activation:}
 \begin{itemize}
   \item \textbf{Caspase-9 activation}: Apoptosome formation
   \item \textbf{Caspase-3/7 activation}: Execution phase initiation
   \item \textbf{PARP cleavage}: DNA repair pathway inactivation
   \item \textbf{Timeline}: Detectable at 24--48 hours, maximal at 72--96 hours
 \end{itemize}
 
 \#\#\# Step 8: Phenotypic Resistance Mechanisms
 \textbf{Relationship: CORRELATIVE (long-term adaptation patterns)}
 
 Prolonged treatment selects for resistant subpopulations:
 
 \textbf{Genetic adaptations:}
 \begin{itemize}
   \item \textbf{KRAS/NRAS mutations}: Bypass BRAF dependency (15--20\% of cases)
   \item \textbf{MEK1/2 mutations}: C121S, P124L/S mutations reduce drug binding
   \item \textbf{RTK amplifications}: EGFR, PDGFR, MET overexpression
 \end{itemize}
 
 \textbf{Epigenetic changes:}
 \begin{itemize}
   \item \textbf{EMT activation}: ZEB1, SNAIL upregulation promoting mesenchymal state
   \item \textbf{Stemness programs}: SOX2, NANOG expression in subpopulations
   \item \textbf{Metabolic reprogramming}: Enhanced oxidative phosphorylation capacity
 \end{itemize}
 \vspace{\baselineskip}
 
 \#\# 4. Measurable Biological Endpoints
 
 \#\#\# Transcriptomic Signatures (detectable by RNA-seq):
 
 \textbf{Early response genes (2--6 hours):}
 \begin{itemize}
   \item \textbf{Downregulated}: MYC, FOS, JUN, EGR1, CCND1, CDK4, CDK6
   \item \textbf{Upregulated}: CDKN1A (p21), GADD45A, DDIT3, ATF3
 \end{itemize}
 
 \textbf{Cell cycle signatures (12--24 hours):}
 \begin{itemize}
   \item \textbf{S-phase genes decreased}: MCM2-7, PCNA, RRM1/2, TYMS
   \item \textbf{DNA repair genes}: BRCA1, RAD51, XRCC1 (context-dependent changes)
 \end{itemize}
 
 \textbf{Metabolic gene signatures (6--24 hours):}
 \begin{itemize}
   \item \textbf{Glycolytic genes down}: SLC2A1 (GLUT1), HK2, PKM2, LDHA
   \item \textbf{Oxidative metabolism}: Complex I-V subunits, PGC1$\alpha$ (variable)
 \end{itemize}
 
 \textbf{Apoptotic signatures (24--72 hours):}
 \begin{itemize}
   \item \textbf{Pro-apoptotic}: BBC3 (PUMA), BCL2L11 (BIM), BAX, BAK1
   \item \textbf{Anti-apoptotic decreased}: MCL1, BCL2L1 (BCL-xL)
 \end{itemize}
 
 \#\#\# Phenomic Signatures (detectable by high-content imaging):
 
 \textbf{Morphological changes:}
 \begin{itemize}
   \item \textbf{Cell size reduction}: 20--30\% decrease in cell area (12--24 hours)
   \item \textbf{Nuclear morphology}: Condensed chromatin, irregular nuclear shape
   \item \textbf{Cytoplasmic changes}: Reduced cytoplasmic volume, organelle reorganization
 \end{itemize}
 
 \textbf{Functional measurements:}
 \begin{itemize}
   \item \textbf{Proliferation rate}: 60--80\% reduction in EdU incorporation (24 hours)
   \item \textbf{Mitochondrial potential}: 30--50\% reduction in TMRM fluorescence (48 hours)
   \item \textbf{Caspase activity}: 5--10 fold increase in caspase-3/7 activity (48--72 hours)
   \item \textbf{Cell viability}: 40--70\% reduction by 96 hours (concentration-dependent)
 \end{itemize}
 
 \textbf{Cell cycle distribution:}
 \begin{itemize}
   \item \textbf{G1 accumulation}: 60--80\% of cells (vs.\ 40--50\% control)
   \item \textbf{S-phase reduction}: 5--15\% of cells (vs.\ 30--40\% control)
   \item \textbf{Sub-G1 population}: 10--30\% at 72--96 hours (apoptotic cells)
 \end{itemize}
 \vspace{\baselineskip}
 
 \#\# 5. Final Phenotypic Outcome
 
 \textbf{Binimetinib treatment induces a cytostatic-to-cytotoxic response in C32 melanoma cells.} The perturbation initially causes \textbf{G1/S cell cycle arrest} through MAPK pathway blockade, leading to \textbf{reduced proliferation} and \textbf{metabolic stress}. Over time, inadequate survival signaling overcomes the hyperactive PI3K/AKT compensation, resulting in \textbf{mitochondrial dysfunction} and \textbf{caspase-dependent apoptosis}.
 
 \textbf{Key measurable outcomes:}
 \begin{itemize}
   \item \textbf{Transcriptomic}: Immediate early gene suppression, cell cycle gene downregulation, delayed apoptotic gene activation
   \item \textbf{Phenomic}: Progressive cell cycle arrest, reduced cell size, decreased mitochondrial function, increased caspase activity, and ultimate cell death
 \end{itemize}
 
 The response demonstrates how MEK inhibition can overcome oncogene addiction in BRAF-mutant melanoma despite multiple resistance mechanisms, making it a clinically effective therapeutic strategy when combined with BRAF inhibitors.
  \end{tcolorbox}

\newpage
\section{Experimental details}\label{appx: exp}

We provide the \Methodname code in {\href{https://github.com/yunhuijang/VC-TRACES}{https://github.com/yunhuijang/VC-TRACES}}.

\subsection{Detailed experimental results}

Here, we present the detailed numerical results of the experiments discussed in \cref{sec5_3:tahoeqa}. We provide the results in \cref{tab_5_2:perturbqa}.

\begin{table*}[h]
    \centering
    \caption{\textbf{TahoeQA performance with total model.} The best results are highlighted in \textbf{bold} and the improvement compared to the vanilla SFT is highlighted with \textcolor{teal}{teal}.}\label{tab_5_2:perturbqa}
    \resizebox{0.85\linewidth}{!}{
    \begin{tabular}{ccccccccc} 
    \toprule[1.25pt]
    Task & Model & C32 & HOP62 & HepG2/C3A & Hs 766T & PANC-1 & {Average} & Union \\ 
    \midrule
    \multirow{9}{*}{Differential expression} 
    & Random & 0.269 & 0.254 & 0.232 & 0.239 & 0.301 & {0.259} & 0.286 \\
    & Mean & 0.405 & 0.419 & 0.345 & 0.465 & 0.330 & {0.393} & 0.401 \\
    & K-neighbor & 0.459 & 0.419 & 0.343 & 0.453 & 0.355 & {0.406} & 0.403 \\
    & STATE (ST) & 0.273 & 0.280 & 0.199 & 0.219 & 0.313 & {0.257} & 0.251 \\
    \cmidrule(lr){2-9} 
    & Zero-shot & 0.363 & 0.423 & 0.316 & 0.332 & 0.399 & {0.367} & 0.338 \\
    \cmidrule(lr){2-9} 
    & SFT & 0.344 & 0.353 & 0.163 & 0.363 & 0.236 & {0.292} & 0.291 \\
    & Ours (SFT - Prompt) & \textcolor{teal}{\textbf{0.470}} & \textcolor{teal}{\textbf{0.470}} & \textcolor{teal}{\textbf{0.362}} & \textcolor{teal}{\textbf{0.470}} & \textcolor{teal}{\textbf{0.405}} & \textcolor{teal}{\textbf{0.435}} & \textcolor{teal}{\textbf{0.452}} \\ 
    & Ours (SFT - Generate) & \textcolor{teal}{0.412} & \textcolor{teal}{0.446} & \textcolor{teal}{0.328} & \textcolor{teal}{0.424} & \textcolor{teal}{0.329} & \textcolor{teal}{{0.388}} & \textcolor{teal}{0.441} \\ 
    \midrule
    \multirow{9}{*}{Direction of change} 
    & Random & 0.525 & 0.526 & 0.517 & 0.522 & 0.558 & {0.530} & 0.513 \\
    & Mean & 0.847 & 0.813 & 0.802 & 0.812 & 0.811 & {0.817} & 0.810 \\
    & K-neighbor & 0.798 & 0.775 & 0.761 & 0.736 & 0.784 & {0.771} & 0.757 \\
    & STATE (ST) & 0.756 & 0.685 & 0.704&0.737 & 0.721 & {0.721} & 0.695 \\
    \cmidrule(lr){2-9} 
    & Zero-shot & 0.101 & 0.063 & 0.055 & 0.049 & 0.061 & {0.066} & 0.062 \\
    \cmidrule(lr){2-9} 
    & SFT & 0.839 & 0.810 & 0.815 & \textbf{0.822} & {0.827} & {0.823} & \textbf{0.820} \\ 
    & Ours (SFT - Prompt) & \textcolor{teal}{\textbf{0.855}} & \textcolor{teal}{\textbf{0.827}} & \textcolor{teal}{\textbf{0.830}} & 0.818 & \textcolor{teal}{\textbf{0.832}} & \textcolor{teal}{\textbf{0.832}} & \textcolor{teal}{\textbf{0.820}} \\ 
    & Ours (SFT - Generate) & 0.702 & 0.707 & 0.704 & 0.688 & 0.773 & {0.715} & 0.727 \\ 
    \bottomrule[1.25pt]
    \end{tabular}
    }
    
\end{table*}

\subsection{Prompts}

\paragraph{Report generator.} This is the prompt used for report generator.

  \begin{tcolorbox}[
    colback=gray!5!white, 
    colframe=gray!60!black,
    title={
      \parbox[t]{\dimexpr\linewidth-4mm\relax}{%
        \ttfamily
        Prompt for report generator
      }
    },
    breakable
  ]
  \# Biomedical Reasoning Assistant – Report Generation

\vspace{\baselineskip}
You are a biomedical reasoning assistant.

Your task is to generate a **comprehensive mechanistic report** describing how the specified perturbation affects cellular biology in the given context.  
The report will later be transformed into a structured explanation, so include **all details necessary for reasoning** that links the perturbation to the phenotype.

\vspace{\baselineskip}

---

\#\# INSTRUCTIONS

1. **Describe the perturbation in detail**  

   - Include its type (chemical, genetic, etc.), primary target(s), known binding affinities or potencies, and mechanism of action (e.g., ATP-competitive inhibition, PROTAC-mediated degradation).

2. **Map the full causal chain step by step**  

   - For example, you can **start from the perturbation** → molecular target → pathway modulation → downstream molecular changes → **phenotypic effect**.  
   
   - Explicitly mark whether each step is:
   
     - **Causal** (direct mechanistic or experimental evidence)  
     
     - **Correlative** (statistical or inferred association)

3. **Prioritize measurable end nodes (effects)** 

   - The **final nodes in the chain should, whenever possible, correspond to (or can be inferred from) measurable outputs from the assays available**:  
   
     - **Transcriptomics**: changes in individual gene expression or gene signatures.  
     
     - **Phenomics (imaging/physiology)**: morphological or functional features (e.g., cell size, nuclear morphology, proliferation rate, apoptosis, mitochondrial potential).  
     
   - **Only use higher-order functional or disease phenotypes** if these can be clearly linked to transcriptomic or phenomic signatures.

4. **Include associative evidence** and **ontological evidence** when available

   - Add correlations, transcriptomic signature similarities, or phenotypic fingerprint associations when direct causality is unclear.

5. **Summarize the final phenotypic outcome**  

   - Explicitly state whether the perturbation induces, rescues, or exacerbates the measured phenotype and what the phenotype is.

6. **Provide quantitative and qualitative details when you have them**

   - Affinities (IC50, Kd), phosphorylation sites, key genes modulated, direction of regulation, morphological metric shifts, etc.

The report should be sufficiently detailed to reconstruct the full reasoning path from **perturbation --> measurable biological effect** 
and can be used to generate hypotheses with all the details for their falsification.

\vspace{\baselineskip}

\#\# QUESTION

**Q: How does the following perturbation influence the cell in the described context, mechanistically and functionally?**  

\{treatment\}
\vspace{\baselineskip}

\#\#\# EXPECTED OUTPUT

Write a **detailed mechanistic report** explaining step by step how this perturbation affects the biology in this context, ending with the **measurable transcriptomic or phenomic effects whenever possible**.

  \end{tcolorbox}

\paragraph{Explanation constructor.} This is the prompt used for explanation constructor.

\begin{tcolorbox}[
    colback=gray!5!white, 
    colframe=gray!60!black,
    title={
      \parbox[t]{\dimexpr\linewidth-4mm\relax}{%
        \ttfamily
        Prompt for explanation constructor
      }
    },
    breakable
]

\# Biomedical Reasoning Assistant - Structured Explanation Task
\vspace{\baselineskip}

You are a biomedical reasoning assistant. Explain how a specific perturbation alters cellular biology in a given context using both causal mechanisms and associative evidence. Your goal is to translate a mechanistic report into a structured, falsifiable explanation of the perturbation's effect.

\vspace{\baselineskip}

\#\# Task Overview
\vspace{\baselineskip}

You must produce **four outputs**:

1. **Private Reasoning** (\texttt{<}think\texttt{>})
2. **Mechanism-of-Action Summary** (\texttt{<}answer\texttt{>})
3. **Structured Explanation** (\texttt{<}explain\texttt{>})
4. **Causal DAG of Events** (\texttt{<}dag\texttt{>})

Each step is strictly defined below.

\vspace{\baselineskip}

\#\# 1. Private Reasoning

Wrap your step-by-step biological reasoning inside \texttt{<}think\texttt{>}...\texttt{<}/think\texttt{>}.

- Proceed as if you are discovering the answer for the first time.
- Simulate the biology: consider binding events, cascades, interactions, and regulatory outcomes, but leverage the reports you have been given to guide your reasoning.
- Mark each connection as either causal or correlative.
- Indicate when evidence is inferred, indirect, or weak.

---

\#\# 2. Summary Answer

Wrap a **1-paragraph summary** inside \texttt{<}answer\texttt{>}...\texttt{<}/answer\texttt{>}.

- Clearly describe the mechanism of action (MoA).
- Use proper biological terminology.
- Distinguish between mechanistic (causal) and associative (correlative) findings.
- Conclude with the predicted phenotypic or functional outcome.

---

\#\# 3. Structured Mechanistic Explanation

Wrap your structured actions inside \texttt{<}explain\texttt{>}...\texttt{<}/explain\texttt{>}.

- Begin with 'set\_context(...)' to define background conditions, including cell type/line, disease model, and any prior perturbation.
- Then, \textbf{list one primitive per line}, using only \textbf{allowed action primitives}.
- Each line must follow the \textbf{signature} of the primitive action.
- Each line must include a unique 'id' (e.g., 'id="n3"') for later referencing in the DAG.
- Your last 'explain' step must be a measurable biological output (omics, phenomics, etc.): 'induces\_phenotype(...)', 'alleviates\_phenotype(...)', 'regulates\_expression(...)', 'regulates\_translation(...)'
- Do not end on upstream mechanistic nodes and always ground the explanation in a testable, measurable biological output.
- Do not use the same 'id' for multiple primitives.
- Use the 'via' as optional field to reference upstream mechanisms for causal steps or explain the evidence basis for correlative ones.

Example line:

binds\_to(id="n1", actor="EW-7197", target="TGFBR1", affinity="0.010 µM", via="ATP site occupancy")

---

\#\# 4. Causal DAG

Output a directed acyclic graph (DAG) inside \texttt{<}dag\texttt{>}...\texttt{<}/dag\texttt{>}.

- The DAG must reference the 'id's declared in the primitives in the \texttt{<}explain\texttt{>} block.
- Define directionality (edge types)
  - 'causal': direct mechanistic consequence
  - 'correlative': statistical or inferred association without mechanistic support
  
For example:
edge("source\_id", "target\_id", relation="causal|correlative")

---

\#\# Allowed Primitives

Use only the following set of structured primitives:

\{action\_primitives\}

Each primitive has a specific signature and biological scope. Refer to the full list above for descriptions and valid fields.

---

\#\# Final Output Format

\texttt{<}think\texttt{>}

...step-by-step causal and correlative reasoning...

\texttt{<}/think\texttt{>}

\texttt{<}answer\texttt{>}

...Mechanism-of-action and functional effect summary...

\texttt{<}/answer\texttt{>}

\texttt{<}explain\texttt{>}

  set\_context(cell\_type="...", genotype="...", disease="...", prior\_perturbation="...")
  primitive\_1(...)
  primitive\_2(...)
  ...
  
\texttt{<}/explain\texttt{>}

\texttt{<}dag\texttt{>}

edge("n1", "n2", relation="causal")
edge("n2", "n3", relation="causal")
edge("n3", "n4", relation="correlative")
...

\texttt{<}/dag\texttt{>}

\vspace{\baselineskip}
\#\#\# Example

Q: How does the following perturbation influence the cell in the described context, mechanistically and functionally?

\{
    "context": \{
        "perturbation\_type": "loss-of-function",
        "description": "CRISPR knockdown of the LEMD3 gene",
        "cell\_type": "N/A",
        "disease\_model": "Buschke-Ollendorff syndrome"
    \},
    
    "perturbation": \{
        "type": "chemical",
        "smiles": "CC1=NC(=CC=C1)C2=C(N=C(N2)CNC3=CC=CC=C3F)C4=CN5C
        (=NC=N5)C=C4",
        "name": "EW-7197",
        "target": "ALK5",
        "moa\_type": "inhibitor"
    \}  
\}

A:
\texttt{<}think\texttt{>}

LEMD3 LoF causes hyper-TGF-$\beta$ signalling (causal).
EW-7197 is an ATP-competitive inhibitor of TGFBR1 (causal).
Binding $\rightarrow$ TGFBR1 activity $\downarrow$ (causal) $\rightarrow$ SMAD2/3 phosphorylation $\downarrow$ (causal).
Lower p-SMAD2/3 $\rightarrow$ nuclear import $\downarrow$ $\rightarrow$ ECM genes COL1A1 \& FN1 $\downarrow$ (causal).
High COL1A1/FN1 correlates with fibrotic matrix overproduction (correlative).
Therefore EW-7197 is likely to rescue the fibrotic phenotype.

\texttt{<}/think\texttt{>}

\texttt{<}answer\texttt{>}

EW-7197 binds the TGF-$\beta$ type I receptor (TGFBR1) with low-nanomolar affinity, competitively inhibiting its kinase activity. Reduced TGFBR1 signaling lowers SMAD2/3 phosphorylation and nuclear translocation, leading to
down-regulation of collagen-I (COL1A1) and fibronectin (FN1) transcripts. Because elevated COL1A1/FN1 expression is
associated with excessive extracellular-matrix deposition in LEMD3-deficient fibroblasts, EW-7197 is predicted to rescue the fibrotic-matrix phenotype characteristic of Buschke-Ollendorff syndrome.

\texttt{<}/answer\texttt{>}

\texttt{<}explain\texttt{>}

  set\_context(cell\_type="dermal fibroblast", disease\_context="Buschke-Ollendorff syndrome", prior\_perturbation="LEMD3 CRISPR knockdown")
  binds\_to(id="n1", actor="EW-7197", target="TGFBR1", affinity="0.010 µM", via="ATP site occupancy")
  modulates\_molecule\_activity(id="n2", target="TGFBR1", direction="down", via="competitive inhibition")
  modulates\_molecule\_activity(id="n3", target="SMAD2/3 complex", direction="down", via="reduced phosphorylation by TGF$\beta$ receptor")
  regulates\_expression(id="n4", regulator="SMAD2/3 complex", gene\_list=["COL1A1","FN1"], direction="down", via="loss of SMAD2/3 nuclear translocation")
  alleviates\_phenotype(id="n5", actor="EW-7197", phenotype="fibrotic matrix overproduction", via="lower ECM gene expression")
  
\texttt{<}/explain\texttt{>}

\texttt{<}dag\texttt{>}

edge("n1", "n2", relation="causal")
edge("n2", "n3", relation="causal")
edge("n3", "n4", relation="causal")
edge("n4", "n5", relation="correlative")
\texttt{<}/dag\texttt{>}

\vspace{\baselineskip}

\# Question

\{question\}
\vspace{\baselineskip}

\# Original Report

\{report\}

\end{tcolorbox}

\paragraph{LLM-judge.} This is the prompt used for LLM-judge.

  \begin{tcolorbox}[
    colback=gray!5!white, 
    colframe=gray!60!black,
    title={
      \parbox[t]{\dimexpr\linewidth-4mm\relax}{%
        \ttfamily
        Prompt for LLM-judge
      }
    },
    breakable
  ]
    You are an expert biologist. Evaluate the following response for its scientific plausibility, coherence and correctness.
\vspace{\baselineskip}

    Question:
    
    \{question\}
    
\vspace{\baselineskip}
    Response:
    
    \{response\}
    \vspace{\baselineskip}

    Assess the explanation on a scale of 0-10 for the following criteria:

    \vspace{\baselineskip}
    1. **"scientific\_accuracy"**:
    
        * **Description**: Are the biological claims, pathways, and interactions factually correct according to current scientific consensus? Are gene/protein names correct? Penalize assertions with low confidence or known inaccuracies.
        
        * **Score**: [0-10]
        
        * **Instruction**: "confidence="low"", "confidence="lost"" should be penalized, the trace should have lower score (0-3), "confidence="correlated"" and "confidence="inferred"" should also be be penalized where the trace should have score 4-5.
        Unless all steps are fully specified, it shouldn't output a score between 8-10.
        \vspace{\baselineskip}
        
    2.  **"logical\_consistency"**:
    
        * **Description**: Does the explanation present a coherent, logical argument? Do the conclusions drawn logically follow from the premises provided within the text?
        
        * **Score**: [0-10]
        
        * **Instruction**: 
            * If there is a loss of function of gene x, it would be wrong if any of the following trace has "binds\_to" to x protein. The "logical\_consistency" should be penalized.
            * If the target of "binds\_to" is "unknown cellular targets", the following traces would be meaningless. The "logical\_consistency" should be penalized.
            Unless all steps are fully specified, it shouldn't output a score between 8-10.
            \vspace{\baselineskip}
            
    3. **"mechanistic\_clarity"**:
    
        * **Description**: How clearly is the underlying biological mechanism explained? Vague or ambiguous terms should be penalized. 
            Penalize missing actors/targets, unspecified directions, hand-wavy pathways, unlabeled compartments.
            
        * **Score**: [0-10]
        
        * **Instruction**: 
        
            * "binds\_to": Penalize missing actors/targets. "actor" and "target" should be specific (e.g. "a protein" is not a specified term for proteins and should be penalized. A compound should be specified by name such as "aspirin" or a SMILES syntax. A term like "unknown" or "a drug" should be heavily penalized.).
            
            * "regulates\_expression": "gene\_list" should be a specific list of genes (e.g., ["FASN","ACC1","SCD1"]) or a well known gene group (e.g., interferon-stimulated genes in interferon signaling pathway). The "direction" must be specified (e.g., "up" or "upregulate", "down" or "downregulate") and terms like "unknown" should be heavily penalized.
            
            * "induces\_phenotype"/"alleviates\_phenotype": "phenotype" must be specific (e.g., "apoptosis," not "a cellular process").
            
            * "localizes\_to": The "entity" must be a specific molecule and the "to\_loc" must be a specific compartments (e.g., "nucleus," not "part of the cell").
            
            * "modulates\_molecule\_activity": The "target" has to be a specified molecule or protein. Ambiguous terms like 'cellular proteins' or 'a molecule' should be heavily penalized.
            
            * "modulates\_pathway\_activity": The "pathway" has to be a specified and well defined pathway (e.g., MAPK pathway). Any hand-wavy pathway name (e.g., cellular process) should be penalized.
            
            Unless all steps are fully specified, testable, clear, it shouldn't output a score between 8-10.

    Your response MUST be a single JSON object containing the numeric scores (0-10) for each criterion. Do not include any other text or explanations.
    You should only use the following keys: "scientific\_accuracy", "logical\_consistency", "mechanistic\_clarity".
    Example of a valid response:
    \{\{"scientific\_accuracy": 8, "logical\_consistency": 9, "mechanistic\_clarity": 7\}\}
  \end{tcolorbox}

\subsection{Hyperparameters}

We report the detailed hyperparameters used for the supervised fine-tuning (SFT) and the subsequent inference generation in \cref{sec5_3:tahoeqa}.

\paragraph{Training Hyperparameters.} The model was fine-tuned with a learning rate of $2×10^{-4}$ using a linear scheduler and a warmup ratio of 0.05. To stabilize training, we employed a weight decay of 0.01 and set the gradient accumulation steps to 4. For parameter-efficient fine-tuning, we utilized LoRA with a rank ($r$) of 64. All experiments were conducted with a fixed random seed of 11 to ensure reproducibility.

\paragraph{Inference Hyperparameters.} For generating structured explanations during evaluation and downstream tasks, we adopted specific sampling strategies to balance diversity and precision. We set the temperature to 0.2, combined with nucleus sampling (top-p) of 0.8 and a top-k value of 20.

\subsection{Computational resources}

We used H100 GPUs for TahoeQA task described in \cref{sec5_3:tahoeqa}.

\newpage

{\section{Additional experiments}\label{appx:additional_exp}}

\subsection{Ablation study: Retrieval source}

We ablate the contribution of each retrieval source by generating explanation traces using only a single knowledge base at a time, compared to the full combination used by \Methodname. We employ the same metrics as in \cref{tab_5_1:main}. As shown in \cref{tab_appx_4_1:retrieval}, no single source approaches the performance of the full combination. This confirms that each database contributes complementary biological knowledge, with StarkPrimeKG providing relational context, Harmonizome enriching gene-level information, PubMed supplying literature-grounded evidence, and Wikipedia offering broad background knowledge.

\begin{table*}[h]
    \centering
    \caption{\textbf{Ablation study on retrieval source.} The best results are highlighted in \textbf{bold}. The standard deviation is computed across cell lines.}\label{tab_appx_4_1:retrieval}
    \resizebox{0.85\linewidth}{!}{
    \begin{tabular}{ccccccc}
    \toprule[1.25pt]
    & \multicolumn{2}{c}{Format} & \multicolumn{2}{c}{Verifier} \\
        \cmidrule(lr){2-3}  \cmidrule(lr){4-5}
        \textbf{Retrieval source} & Validity & Verifiability &  Drug-target interaction & Differential expression \\
        \midrule
         PubMed & 1.000 \scriptsize{$\pm$ 0.000} & 0.300 \scriptsize{$\pm$ 0.099} & 0.432 \scriptsize{$\pm$ 0.059} & 0.435 \scriptsize{$\pm$ 0.050} \\
         StarkPrimeKG & 1.000 \scriptsize{$\pm$ 0.000} & 0.296 \scriptsize{$\pm$ 0.089} & 0.382 \scriptsize{$\pm$ 0.031} & 0.449 \scriptsize{$\pm$ 0.054} \\
         Harmonizome & 1.000 \scriptsize{$\pm$ 0.000} & 0.297 \scriptsize{$\pm$ 0.092} & 0.399 \scriptsize{$\pm$ 0.013} & 0.457 \scriptsize{$\pm$ 0.051} \\
         Wikipedia & 1.000 \scriptsize{$\pm$ 0.000} & 0.292 \scriptsize{$\pm$ 0.091} & 0.469 \scriptsize{$\pm$ 0.028} & 0.485 \scriptsize{$\pm$ 0.063} \\
         \midrule
         All (Ours) & \textbf{1.000} \scriptsize{$\pm$ 0.000} & \textbf{0.601} \scriptsize{$\pm$ 0.009} & \textbf{0.766} \scriptsize{$\pm$ 0.008} & \textbf{0.487} \scriptsize{$\pm$ 0.054} \\
     \bottomrule[1.25pt]
    \end{tabular}}
\end{table*}

\subsection{Ablation study: Backbone LLM}

We compare the backbone LLM used in \Methodname (Claude) against GPT-4.1. and Gemini-2.5-Flash, keeping the retrieval pipeline and explanation constructor prompts identical. As shown in \cref{tab_appx_4_2:backbone}, Claude achieves perfect trace validity. On DTI score, Claude substantially outperforms both alternatives. GPT-4.1 achieves a higher DE score, but at the cost of lower verifiability, meaning a large fraction of its outputs are structurally invalid and cannot be reliably evaluated. These results validate Claude as the backbone choice, balancing format adherence with biological accuracy.

\begin{table*}[h]
    \centering
    \caption{\textbf{Ablation study on backbone LLM.} The best results are highlighted in \textbf{bold}. The standard deviation is computed across cell lines.}\label{tab_appx_4_2:backbone}
    \resizebox{0.85\linewidth}{!}{
    \begin{tabular}{ccccccc}
    \toprule[1.25pt]
    & \multicolumn{2}{c}{Format} & \multicolumn{2}{c}{Verifier} \\
        \cmidrule(lr){2-3}  \cmidrule(lr){4-5}
        \textbf{Backbone} & Validity & Verifiability &  Drug-target interaction & Differential expression \\
        \midrule
GPT-4.1 & 0.970 \scriptsize{$\pm$ 0.002} & 0.299 \scriptsize{$\pm$ 0.105} & 0.399 \scriptsize{$\pm$ 0.063} & 0.527 \scriptsize{$\pm$ 0.076} \\
         Gemini-2.5-Flash & 0.060 \scriptsize{$\pm$ 0.016} & 0.963 \scriptsize{$\pm$ 0.078} & 0.338 \scriptsize{$\pm$ 0.036} & 0.207 \scriptsize{$\pm$ 0.036} \\
         \midrule
         Claude (Ours) & \textbf{1.000} \scriptsize{$\pm$ 0.000} & \textbf{0.601} \scriptsize{$\pm$ 0.009} & \textbf{0.766} \scriptsize{$\pm$ 0.008} & \textbf{0.487} \scriptsize{$\pm$ 0.054} \\
     \bottomrule[1.25pt]
    \end{tabular}}
\end{table*}

\subsection{Ablation study: One-step generation}

We evaluate whether the two-stage pipeline (report generation followed by explanation construction) is necessary by comparing it against a one-step baseline that directly generates structured explanations from the perturbation input and retrieved information without an intermediate report. As shown in \cref{tab_appx_4_3:one_step}, the two-step pipeline nearly doubles both DTI and DE scores over one-step generation, while also achieving higher verifiability. This confirms that decoupling knowledge synthesis from structured reasoning generation is critical.

\begin{table*}[h]
    \centering
    \caption{\textbf{Ablation study on one-step generation.} The best results are highlighted in \textbf{bold}. The standard deviation is computed across cell lines.}\label{tab_appx_4_3:one_step}
    \resizebox{0.85\linewidth}{!}{
    \begin{tabular}{ccccccc}
    \toprule[1.25pt]
    & \multicolumn{2}{c}{Format} & \multicolumn{2}{c}{Verifier} \\
        \cmidrule(lr){2-3}  \cmidrule(lr){4-5}
        \textbf{Step} & Validity & Verifiability &  Drug-target interaction & Differential expression \\
        \midrule
         One-step & 0.999 \scriptsize{$\pm$ 0.002} & 0.534 \scriptsize{$\pm$ 0.067} & 0.329 \scriptsize{$\pm$ 0.026} & 0.267 \scriptsize{$\pm$ 0.051} \\
         \midrule
         Two-step (Ours) & \textbf{1.000} \scriptsize{$\pm$ 0.000} & \textbf{0.601} \scriptsize{$\pm$ 0.009} & \textbf{0.766} \scriptsize{$\pm$ 0.008} & \textbf{0.487} \scriptsize{$\pm$ 0.054} \\
     \bottomrule[1.25pt]
    \end{tabular}}
\end{table*}

\subsection{Ablation study: Given the same retrieved report}

To disentangle the contribution of our structured reasoning formalism from the retrieval advantage, we evaluate all baseline models when given the same retrieved report generated by our report generator as input. This isolates whether the performance gap in \cref{tab_5_1:main} is driven by access to richer context or by the structured explanation pipeline itself. As shown in \cref{tab_appx_4_4:same_report}, \Methodname maintains the highest validity and verifiability even when all models receive identical retrieved context. Notably, while Qwen3 achieves a higher DE score (0.556 vs.\ 0.457), this comes at the cost of substantially lower verifiability (0.340 vs.\ 1.000), meaning that a large fraction of its outputs are structurally invalid and cannot be reliably evaluated or used downstream. Notably, we clarify that the baseline model selection excluding the retrieval in Table 1 is intentional, as the retrieval pipeline is a core component of our framework.

\begin{table*}[h]
    \centering
    \caption{\textbf{Ablation study given the same retrieved report.} The best results are highlighted in \textbf{bold}. The standard deviation is computed across cell lines.}\label{tab_appx_4_4:same_report}
    \resizebox{0.85\linewidth}{!}{
    \begin{tabular}{ccccccc}
    \toprule[1.25pt]
    & \multicolumn{2}{c}{Format} & \multicolumn{2}{c}{Verifier} \\
        \cmidrule(lr){2-3}  \cmidrule(lr){4-5}
        \textbf{Model} & Validity & Verifiability &  Drug-target interaction & Differential expression \\
        \midrule
DeepSeek & 0.046 \scriptsize{$\pm$ 0.051} & 0.711 \scriptsize{$\pm$ 0.269} & 0.185 \scriptsize{$\pm$ 0.099} & 0.246 \scriptsize{$\pm$ 0.145} \\
Qwen3 & 0.937 \scriptsize{$\pm$ 0.055} & 0.340 \scriptsize{$\pm$ 0.173} & 0.081 \scriptsize{$\pm$ 0.056} & 0.556 \scriptsize{$\pm$ 0.152} \\
Llama3.3 & 0.789 \scriptsize{$\pm$ 0.105} & 0.468 \scriptsize{$\pm$ 0.227} & 0.079 \scriptsize{$\pm$ 0.098} & 0.231 \scriptsize{$\pm$ 0.146} \\
         \midrule
         Two-step (Ours) & \textbf{1.000} \scriptsize{$\pm$ 0.000} & \textbf{0.601} \scriptsize{$\pm$ 0.009} & \textbf{0.766} \scriptsize{$\pm$ 0.008} & \textbf{0.487} \scriptsize{$\pm$ 0.054} \\
     \bottomrule[1.25pt]
    \end{tabular}}
\end{table*}

\subsection{Ablation study: Verifier-based filtering}\label{appx_sub_sec:verifier}

In this section, we conduct a component-wise ablation study to evaluate the effectiveness of the verifier-based filtering pipeline proposed in \cref{sec:4_verifier}. To evaluate, we compare the quality of explanation traces under four distinct conditions: (1) No Filtering (baseline), (2) DTI-verifier only, (3) DE-verifier only, and (4) the Full Pipeline. For this evaluation, we utilize the complete explanation traces derived from 18,950 perturbation-context pairs within the Tahoe-100M~\citep{Zhang2025.02.20.639398} dataset.

\paragraph{Metrics.} We assess the quality of the filtered datasets using three LLM-as-judge evaluation metrics:
\begin{itemize}
    \item Scientific accuracy: Measures whether the mechanistic claims are biologically valid and factually grounded.
    \item Logical consistency: Evaluates whether each step follows a coherent progression without contradictions.
    \item Mechanistic clarity: Assesses whether the underlying biological mechanism is articulated clearly.
\end{itemize}

The detailed prompts are provided in \cref{appx: exp}.

\begin{table}[h]
    \centering
    \caption{\textbf{Ablation study on verifier-based filtering.} The best results are highlighted in \textbf{bold}.}\label{tab_5:ablation}
    \resizebox{0.5\linewidth}{!}{
    \begin{tabular}{cccc}
    \toprule[1.25pt]
    & Scientific accuracy & Logical consistency & Mechanistic clarity \\
    \midrule
    Without verifiers & 0.641 & 0.720 & \textbf{0.725} \\
    \midrule
    +DE & 0.642 & 0.721 & \textbf{0.725}\\
    +DTI & 0.643 & 0.721 & \textbf{0.725}\\
    \midrule
    +DE\&DTI & \textbf{0.645} & \textbf{0.724} & \textbf{0.725} \\
     \bottomrule[1.25pt]
    \end{tabular}}
    
    \vspace{-0.1in}
\end{table}

\paragraph{Results.} As presented in \cref{tab_5:ablation}, verifier-based filtering yields modest but consistent improvements in the overall quality of the explanation traces across all metrics. Notably, while explanations refined via verifier-based filtering is designed to improve the mechanistic clarity and factual accuracy, these gains are not always captured by LLM-based evaluators. This limitation stems from the evaluator's  lack of the specialized regulatory knowledge such as those from Boltz-2~\citep{Passaro2025.06.14.659707} and \mbox{Tahoe-100M}~\citep{Zhang2025.02.20.639398}, which are used by our verifiers. Consequently, the evaluator struggles to distinguish between filtered and unfiltered traces as long as both maintain a high degree of surface-level mechanistic plausibility. 

\subsection{Human evaluation}

To validate that our LLM-based evaluation captures biologically meaningful quality, we conducted a human expert evaluation on a randomized subset of generated explanation traces and measured agreement with LLM-judge scores.
 
\paragraph{Setup.}
We randomly sampled 10 explanation traces from \datasetname, stratified across the five cell lines used in our experiments. Each trace was independently scored by domain experts with backgrounds in molecular biology and pharmacology, using the same three criteria used in \cref{appx_sub_sec:verifier}: \emph{scientific accuracy}, \emph{logical consistency}, and \emph{mechanistic clarity}. In addition, experts provided a binary plausibility judgment (plausible vs.\ implausible) for each trace. Annotators were provided with the perturbation-context pair alongside the generated structured explanation, without access to the LLM-judge scores or verifier outputs.
 
\begin{figure}[h]
    \centering
    \includegraphics[width=\linewidth]{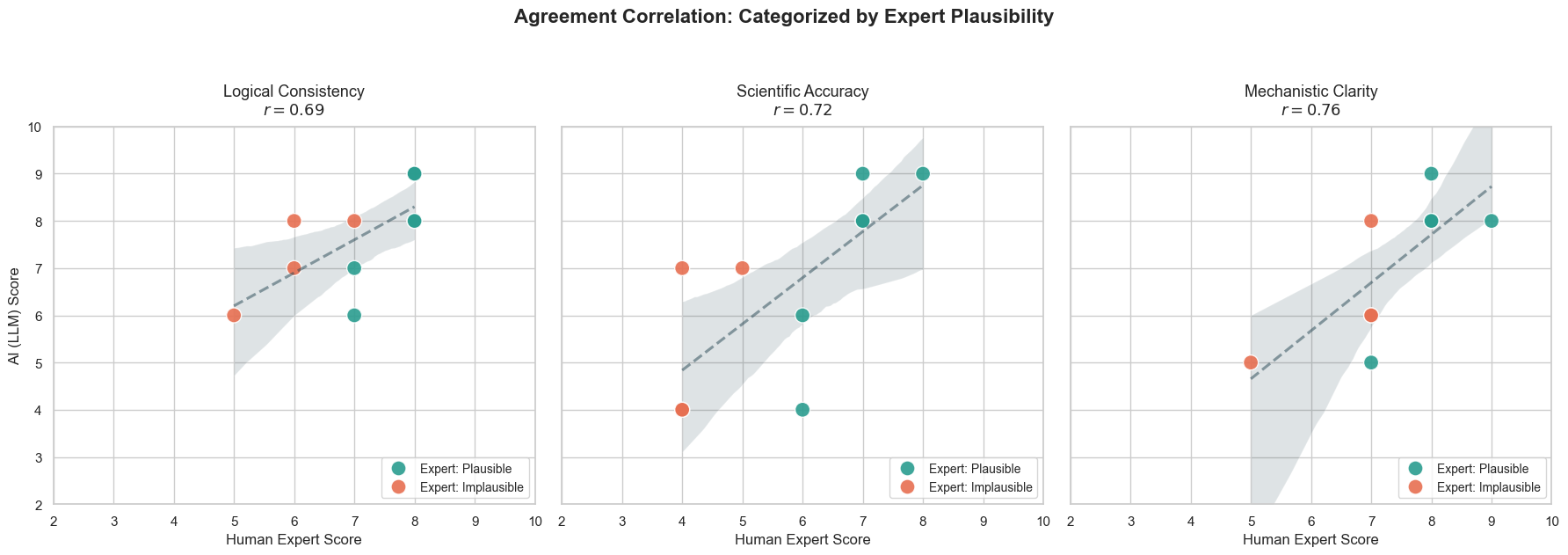}
    \caption{\textbf{Agreement between human expert and LLM-judge scores.} Each point represents a single trace. Teal points indicate traces judged as plausible by experts; orange points indicate implausible traces. Dashed lines show the linear regression fit and shaded regions denote the 95\% confidence interval. Pearson $r$ is reported per criterion.}\label{fig_appx_4_6:human}
\end{figure}

\paragraph{Results.}
We report the Pearson correlation between human expert ratings and LLM-judge scores in \cref{tab_appx_4_6:human} and visualize the per-trace agreement in \cref{fig_appx_4_6:human}. Across all three criteria, we observe strong positive correlation (average $r = 0.72$), confirming that the LLM-judge serves as a reliable proxy for domain expert assessment. We additionally report the average LLM-judge scores across the full set of traces, which indicate consistently high quality across all criteria.

\begin{table}[h]
    \centering
    \caption{\textbf{Human expert evaluation.} Pearson correlation between human expert ratings and LLM-judge scores on a randomized subset of traces, alongside average LLM-judge scores on the full dataset.}\label{tab_appx_4_6:human}
    \resizebox{0.7\linewidth}{!}{
    \begin{tabular}{cccc}
    \toprule[1.25pt]
         & Scientific accuracy & Logical consistency & Mechanistic clarity \\
        \midrule
         Human--LLM correlation ($r$) & 0.72 & 0.69 & 0.76 \\
         Average LLM-judge score (full) & 0.65 & 0.72 & 0.73 \\
     \bottomrule[1.25pt]
    \end{tabular}}
\end{table}

As shown in \cref{fig_appx_4_6:human}, traces judged as plausible by experts (\textcolor{teal}{teal}) consistently cluster in the upper-right region of the score space, while traces judged as implausible (\textcolor{orange}{orange}) tend to receive lower scores from both human experts and the LLM-judge. This alignment indicates that the LLM-judge not only correlates with expert scores numerically but also agrees on which traces are biologically sound, providing confidence that it can serve as a scalable proxy for expert review.

\newpage

{\section{Additional verifiers}\label{appx: additional_verifiers}}

In this section, we provide detailed description of additional verifiers.

\paragraph{LOC verifier.}
The LOC verifier validates the \texttt{localizes\_to} action by cross-referencing the claimed subcellular localization against curated annotations from UniProt~\citep{uniprot} and the Human Protein Atlas~\citep{thul2018human}. For each \texttt{localizes\_to} claim, the verifier checks whether the specified entity is annotated to the claimed \texttt{from\_loc} and \texttt{to\_loc} compartments.
 
\paragraph{PHENO verifier.}
The PHENO verifier assesses the \texttt{induces\_phenotype} and \texttt{alleviates\_phenotype} actions by querying phenotypic databases. Specifically, it maps the perturbation to known phenotypic profiles documented in the Cellular Phenotype Database~\citep{kirsanova2015cellular}. The verifier checks whether the claimed phenotype is consistent with documented phenotypic associations for the perturbation or its downstream targets.

We emphasize that the verification suite is designed to be extensible; as reliable computational tools become available for additional action types, new verifiers can be incorporated into the pipeline without modifying the overall architecture.

\end{document}